\documentclass{article}

\usepackage[preprint,nonatbib]{neurips_data_2024}
\usepackage{natbib}
\setcitestyle{authoryear}

\usepackage[utf8]{inputenc} % allow utf-8 input
\usepackage[T1]{fontenc}    % use 8-bit T1 fonts
\usepackage[backref=page]{hyperref}       % hyperlinks
\usepackage{url}            % simple URL typesetting
\usepackage{booktabs}       % professional-quality tables
\usepackage{amsfonts}       % blackboard math symbols
\usepackage{nicefrac}       % compact symbols for 1/2, etc.
\usepackage{microtype}      % microtypography
\usepackage{xcolor}         % colors
\usepackage{multirow}
\usepackage{graphicx}
\usepackage{setspace}
\usepackage{adjustbox}

\usepackage{caption}
\usepackage{subcaption}         % Allows for the use of subfigures and subcaptions
\usepackage{float}
\definecolor{bluecite}{HTML}{0875b7}
\hypersetup{
  colorlinks=true,
  citecolor=bluecite,
  linkcolor=magenta
}
\usepackage{amsmath}
\usepackage{amssymb}
\usepackage{bm}
\usepackage{makecell}
\usepackage{wrapfig}
\usepackage{enumitem}
\usepackage{pdflscape}

\usepackage{listings}

\definecolor{codegreen}{rgb}{0,0.6,0}
\definecolor{codegray}{rgb}{0.5,0.5,0.5}
\definecolor{codepurple}{rgb}{0.58,0,0.82}
\definecolor{backcolour}{rgb}{0.95,0.95,0.92}
\lstdefinestyle{mystyle}{
    backgroundcolor=\color{backcolour},   
    commentstyle=\color{codegreen},
    keywordstyle=\color{magenta},
    numberstyle=\tiny\color{codegray},
    stringstyle=\color{codepurple},
    basicstyle=\ttfamily\footnotesize,
    breakatwhitespace=false,         
    breaklines=true,                 
    captionpos=b,                    
    keepspaces=true,                 
    numbers=left,                    
    numbersep=5pt,                  
    showspaces=false,                
    showstringspaces=false,
    showtabs=false,                  
    tabsize=2
}
\definecolor{cmarkcolour}{HTML}{219e1c}
\definecolor{xmarkcolour}{HTML}{c4231a}

\newcommand{\cmark}{\color{cmarkcolour}{\large $\checkmark$}}%
\newcommand{\xmark}{\color{xmarkcolour}{$\bullet$}}%        

\usepackage{tcolorbox}
\newtcolorbox{mybox}[2]{float=hp,colback=bluecite!5!white,colframe=bluecite!75!black,fonttitle=\bfseries,title={#1}, label=#2,width=\linewidth,size=small}

\definecolor{plotblue}{HTML}{5799C7}
\definecolor{plotorange}{HTML}{FF9F4A}
\definecolor{plotgreen}{HTML}{61B861}
\definecolor{plotred}{HTML}{E05D5E}
\definecolor{plotpurple}{HTML}{AF8CCD}

\newcommand{\maxresult}[2]{\textcolor{black}{$#1\pm#2$}}
\newcommand{\meanresult}[2]{\textcolor{black}{$\langle#1\pm#2\rangle$}}
\newcommand{\finalresult}[2]{\textcolor{black}{$(#1\pm#2)$}}

\newcommand{\RESULT}[6]{
    \makecell[c]{\maxresult{#1}{#2} \\ \meanresult{#3}{#4} \\ \finalresult{#5}{#6}}
}

\title{Dispelling the Mirage of Progress in Offline MARL through Standardised Baselines and Evaluation}

\author{%
  Claude Formanek$^{1,2}$\thanks{Corresponding author: \texttt{c.formanek@instadeep.com}}
  \And
  Callum Rhys Tilbury$^1$
  \And
  Louise Beyers$^1$
  \AND
  Jonathan Shock$^{2,3,4}$
  \And
  Arnu Pretorius$^{1}$ \and
  \\
  $^{1}$InstaDeep \quad
  $^{2}$University of Cape Town \quad
  $^{3}$INRS, Montreal \quad
  $^{4}$NITheCS, Stellenbosch
}

\begin{document}

\maketitle

\begin{abstract}
Offline multi-agent reinforcement learning (MARL) is an emerging field with great promise for real-world applications. Unfortunately, the current state of research in offline MARL is plagued by inconsistencies in baselines and evaluation protocols, which ultimately makes it difficult to accurately assess progress, trust newly proposed innovations, and allow researchers to easily build upon prior work. In this paper, we firstly identify significant shortcomings in existing methodologies for measuring the performance of novel algorithms through a representative study of published offline MARL work. Secondly, by directly comparing to this prior work, we demonstrate that simple, well-implemented baselines can achieve state-of-the-art (SOTA) results across a wide range of tasks. Specifically, we show that on $35$ out of $47$ datasets used in prior work (almost 75\% of cases), we match or surpass the performance of the current purported SOTA. Strikingly, our baselines often substantially outperform these more sophisticated algorithms.  Finally, we correct for the shortcomings highlighted from this prior work by introducing a straightforward standardised methodology for evaluation and by providing our baseline implementations with statistically robust results across several scenarios, useful for comparisons in future work. Our proposal includes simple and sensible steps that are easy to adopt, which in combination with solid baselines and comparative results, could substantially improve the overall rigour of empirical science in offline MARL moving forward.
\end{abstract}

\section{Introduction}

Offline reinforcement learning (RL) attempts to derive optimal sequential control policies from static data alone, without access to online interactions (e.g.~a simulator). Though the single-agent variant has received fairly widespread research attention~\citep{prudencio2023offlinerl}, progress in the multi-agent context has been slower, due to a variety reasons. For one, multi-agent problems are fundamentally more difficult, and bring a host of new challenges---including difficulties in coordination~\citep{barde2023modelbased}, large joint-action spaces \citep{maicq}, heterogeneous agents \citep{zhong2024heterogeneous} and non-stationarity~\citep{papoudakis2019dealing}, which are absent in single-agent situations.

Nonetheless, some progress has been made in offline multi-agent reinforcement learning (MARL) in recent years. In particular, in better understanding the aforementioned difficulties of offline learning in the multi-agent setting and proposing certain remedies for them \citep{jiang2023offline,maicq,omar,omiga,cfcql,tian2023goodtrajectories,madt}.

% highlight that in offline MARL the transition dynamics in the dataset can significantly differ from those of the learned policies, leading to coordination failures. \cite{} show that the policy gradients in MADDPG~\citep{lowe2017maddpg} and Independent DDPG~\citep{lillicrapContinuousControlDeep2016} struggle to optimise conservative Q-functions (CQL)~\citep{kumar2020cql} and \citet{} further highlight the limitations of scaling MADDPG with CQL, while \cite{} explore a novel approach to regularising the value function in Offline MARL. Finally, \citet{} consider an imbalance of agent expertise in a dataset, which can contaminate the offline learning of all agents. 
% They address this problem by learning decomposed rewards, and then reconstructing the dataset while favouring high-return individual trajectories.
% and propose a per-agent CQL regularisation that scales better in the number of agents.
% They address this by normalising the transition probabilities in the dataset. \cite{maicq} highlight and address the rapid accumulation of extrapolation error due to \textit{Out-of-Distribution} actions in Offline MARL.
% and propose using a zeroth-order optimisation to learn more coordinated behaviour. 

\begin{figure}[t]
     \centering
    \includegraphics[width=1\textwidth]{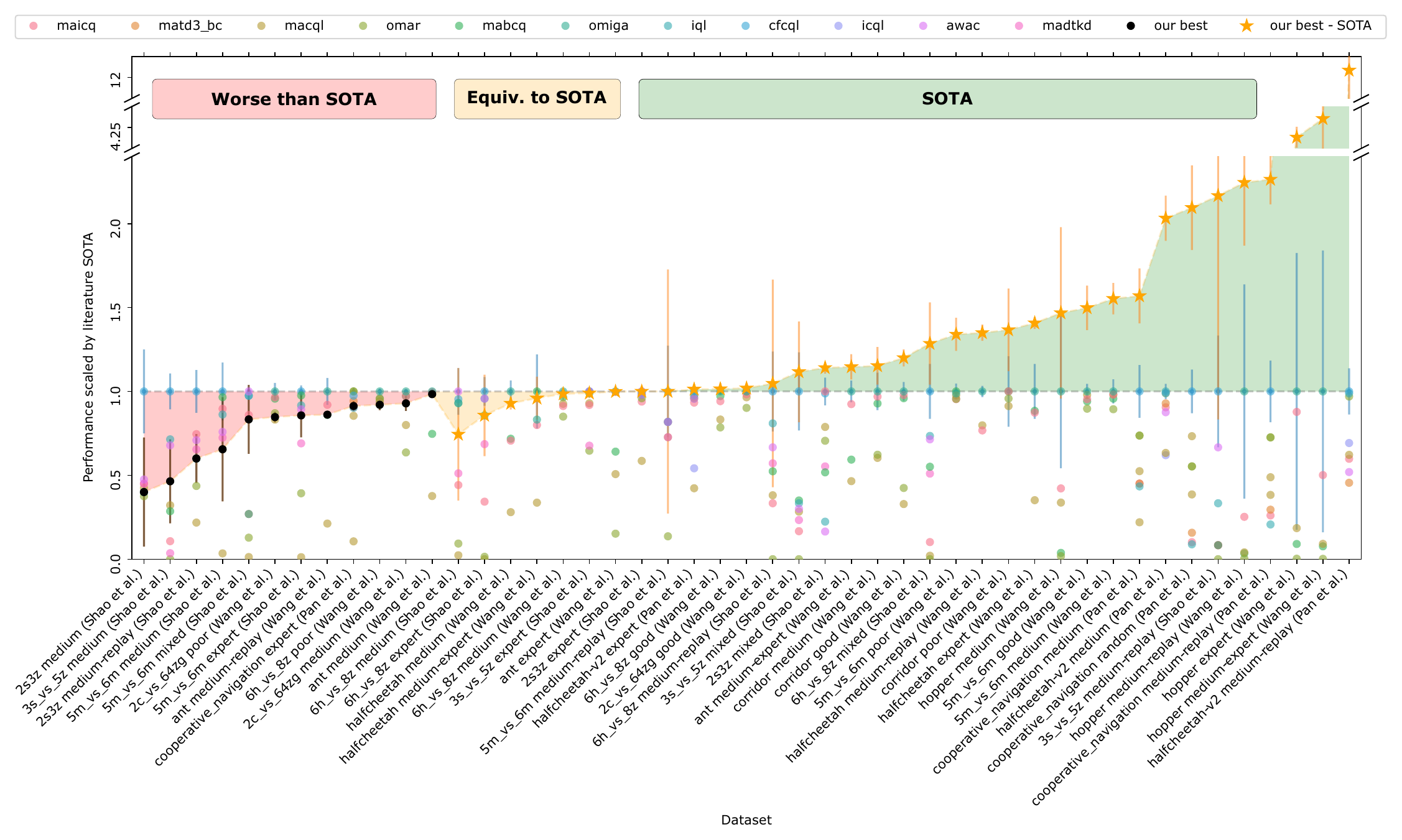}
     \caption{We compare our baseline implementations to the reported performance of various algorithms from the literature across a wide range of datasets. We normalise results from each dataset (i.e.~scenario-quality-source combination) by the SOTA performance from the literature for that dataset. Standard deviation bars are given and when our baseline is significantly better or equal to the best method, using a two-side t-test, we indicate so using a gold star. \textbf{We find that on 35 out of the 47 datasets tested (almost 75\% of cases), we match or surpass the performance of the current SOTA.}}
     \label{fig:comp_results}
\end{figure}

However, we argue that this progress might be a mirage and that offline MARL research is ultimately being held back by a lack of clarity and consistency in baseline implementations and evaluation protocols. To support this claim, we demonstrate in Figure~\ref{fig:comp_results} a surprising but telling result---we show that good implementations of straightforward baseline algorithms can achieve state-of-the-art performance across a wide-range of tasks, beating several published works claiming such a title.  We view our analysis as robust, using datasets and environments with experimental settings that exactly match prior work (see Section \ref{sec:benchmarking}).
%(or match as closely as possible when certain information is not available, discussed in detail in Section \ref{sec:benchmarking}).

This paper proceeds as follows. We first assess the state of the field, diagnosing the key points of friction for progress in offline MARL. Thereafter, we describe in more detail how well-implemented baseline methods perform surprisingly well compared to leading methods from the literature, suggesting that algorithmic progress has not advanced at the rate perhaps previously perceived by the community. In response, we introduce a standardised baseline and evaluation methodology, in an effort to support the field towards being more scientifically rigorous. We hope that researchers will build upon this work, advocating for a cleaner, more reproducible and robust empirical science for offline MARL. 

\section{Methodological Problems in Offline MARL}\label{sec:method_problems}

In this section, we briefly assess the state of offline MARL research by focusing on the baselines and evaluation protocols commonly employed. We consider the following five papers, all published at top-tier venues, for our case study: MAICQ~\citep{maicq}, OMAR~\citep{omar}, MADT~\citep{madt}, CFCQL~\citep{cfcql}, and OMIGA~\citep{omiga}. Given the nascency of the field, we consider these papers to serve as a good representative sample of the current trends in offline MARL research. By looking at this cross-section, we can assess the current methodologies for measuring progress. We present our findings below.

\paragraph{Ambiguity in the Naming of Baseline Algorithms} \label{sec:ambiguous_baselines}
In single-agent RL, the naming of a given algorithm is fairly unambiguous: it is widely understood what core algorithmic steps constitute, e.g.,~DDPG~\citep{lillicrapContinuousControlDeep2016}. Yet, in MARL, algorithms become more complex, since one must specify how multiple agents should learn and interact. Unlike in the single agent case, there is ambiguity when referring to \emph{multi-agent} DDPG---for this might be referring to MADDPG~\citep{lowe2017maddpg}, or independent DDPG agents, or perhaps some other way of interleaving training and/or execution with DDPG forming the base of the algorithm's design. The corresponding impact on the performance of such choices can be significant~\citep{lyu2021contrasting}, and thus it is important to be as explicit as possible.

Clarity in this naming has been lacking in offline MARL literature. For instance, we consider Conservative Q-Learning (CQL)~\citep{kumar2020cql} as a prime example of this problem. As an influential single-agent offline RL algorithm, CQL is critical to consider as a baseline when proposing new work in the field. Whereas the core CQL algorithmic steps are well-established, though, there does not exist a common understanding in the literature of what \emph{multi-agent} CQL is, despite widespread appearance of the abbreviation, \texttt{MACQL}, and its permutations. Consider Table~\ref{tab:macql}, which shows how the same baseline method is purportedly included in each of the papers in our case study, yet the details provided for this algorithm are sparse, often lacking publicly available code, and can vary dramatically across papers.

\begin{table}
\centering
\scriptsize
\setstretch{1.5}
\caption{Demonstration of how papers in our case study are essentially using the same name for Multi-Agent CQL, for markedly different algorithms, often providing only sparse information about their implementations.}\label{tab:macql}
\label{table:macql}
\vspace{1em}
\begin{adjustbox}{max width=\textwidth}
\begin{tabular}{c|ccc}
\textbf{Paper} & \textbf{Name for \texttt{MACQL}} & \textbf{Implementation Details Provided in the Paper} & \textbf{Is the Full Code Available?} \\

\hline 
\citet{maicq}
% &mujocomujoco
% \begin{tabular}{c} 
% \symboldiscrete \\
% \symbolcontinuous
% \end{tabular}
& MA-CQL
 & \begin{tabular}{c} 
Loss function, value-decomposition structure
\end{tabular} &
% \textcolor{xmarkcolour}{
No, only single-agent CQL
% }
\\

\hline
\citet{omar}
% &
% \begin{tabular}{c}
% \symboldiscrete \\
% \symbolcontinuous
% \end{tabular}
& MA-CQL
 & \begin{tabular}{c} 
MADDPG~\citep{lowe2017maddpg}\\(Discrete: + Gumbel Softmax~\citep{jang2016categorical})
\end{tabular} & 
% \textcolor{xmarkcolour}{
Only for continuous settings
% }
\\

\hline 
\citet{madt}
% &
% \begin{tabular}{c} 
% \symboldiscrete \\
% \symbolcontinuous
% \end{tabular}
& CQL-MA
 & \begin{tabular}{c} 
CQL + ``mixing network''
\end{tabular} &
% \textcolor{xmarkcolour}{
No
% }
\\

\hline
\citet{cfcql}
% &
% \begin{tabular}{c} 
% \symboldiscrete \\
% \symbolcontinuous
% \end{tabular}
& MACQL
& \begin{tabular}{c} 
``Naive extension of CQL to multi-agent settings'' + Loss function
\end{tabular} & Yes \\

% \hline
% OMAC
% % &
% % \begin{tabular}{c} 
% % \symboldiscrete \\
% % \symbolcontinuous
% % \end{tabular}
% & CQL-MA
% & \begin{tabular}{c} 
% Value decomposition structure + policy constraint
% \end{tabular} & \textcolor{xmarkcolour}{No} \\

\hline
\citet{omiga}
% &
% \begin{tabular}{c} 
% \symboldiscrete \\
% \symbolcontinuous
% \end{tabular}
& CQL-MA
& \begin{tabular}{c} 
Value decomposition structure + policy constraint
\end{tabular} &
No
\\

\bottomrule
\end{tabular}
\end{adjustbox}
\end{table}

\begin{wrapfigure}{r}{0.6\textwidth}
  \centering
    \includegraphics[width=\linewidth]{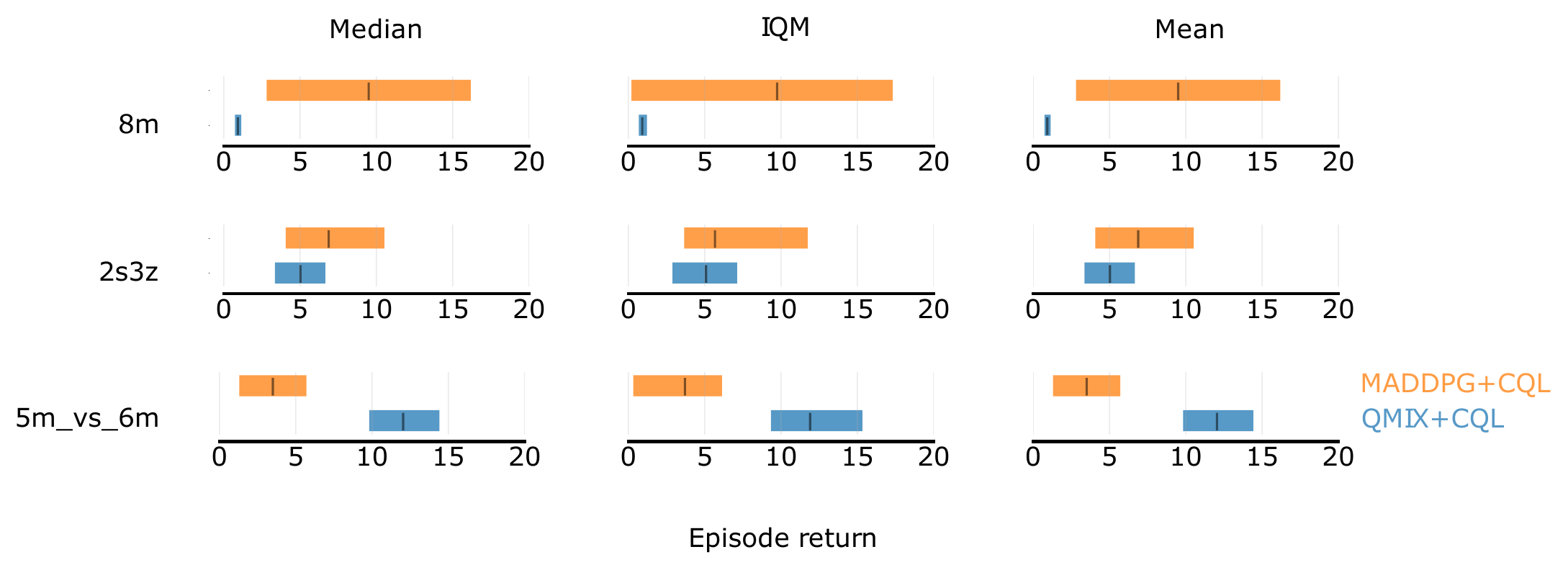}
  \caption{Comparing the performance of QMIX+CQL and \mbox{MADDPG+CQL}, two algorithms that could reasonably be called \texttt{MACQL} in the literature (see Table~\ref{table:macql}), using the Medium dataset from three different SMACv1 scenarios. We see that the difference in performance of these algorithms is significant, and depends on the scenario considered.}
    \label{fig:qmix_maddpg}
\end{wrapfigure}

To highlight these discrepancies and their effects more clearly, we note that the authors of each paper compare their proposed method with ``\texttt{MACQL}'' and claim superior performance. But \emph{which} \texttt{MACQL} is being compared, has a significant bearing on the degree to which these conclusions are likely to hold. For example, consider the following simple experiment. We compare two viable candidates for \texttt{MACQL} across three SMACv1~\citep{smac} maps (\texttt{8m}, \texttt{2s3z}, \texttt{5m v 6m}). Specifically, we compare MADDPG~\citep{lowe2017maddpg} with the Gumbel-Softmax~\citep{jang2016categorical}, against QMIX~\citep{rashid2018qmix}, each with the addition of CQL. Importantly, note that in both cases, we could present such an algorithm as ``\texttt{MACQL}'' as used in prior work. However, the results in Figure~\ref{fig:qmix_maddpg} clearly reveal their relative differences in performance. Here we report the median, mean and interquartile mean (IQM) as recommended by \citet{agarwal2022deep}. We notice, too, that the outcome changes depending on the scenario used---MADDPG with CQL outperforms QMIX on \texttt{8m}, whereas the ordering is reversed on \texttt{5m\_vs\_6m}.

\paragraph{Mismatches in Scenarios Used for Comparison}
In addition to the ambiguity in baselines used for comparison, there is also a confusing mismatch in the selected scenarios used across prior work. Quite simply, consider Table~\ref{tab:baseline_scenarios}, which shows the choice of scenarios from SMACv1 used for comparison across the different papers in our case study. There is not a single scenario that is used consistently in all papers. Furthermore, we find several instances where a scenario is unique to a specific paper, even though in practice, it should be trivial to ensure an overlap with all scenarios from prior work. As a result, it becomes difficult to trust and corroborate the results across different papers, and make meaningful comparisons when a new algorithm is proposed.

\begin{table}
\centering
\scriptsize
\setstretch{1.5}
\caption{Depiction of which SMACv1 scenarios were used in the five papers from our case study\\({\cmark \, present}, {\xmark \, absent}).}
\vspace{1em}
\label{tab:baseline_scenarios}
\begin{adjustbox}{max width=\textwidth}
% \begin{tabular}{c|c|c|c|c|c|c|c|c|c|c|c|c}
\begin{tabular}{c|cccccccccccc}
\textbf{Paper} & \texttt{2s3z} & \texttt{3s v 5z} & \texttt{5m v 6m} & \texttt{6h v 8z} & \texttt{MMM} & \texttt{10m v 11m} & \texttt{3s v 3z} & \texttt{3s5z} & \texttt{1c3s5z} & \texttt{2c v 64zg} & \texttt{3s5z v 3s6z} & \texttt{corridor} \\ \hline
 \citet{maicq} & \cmark & \xmark & \xmark & \xmark & \cmark & \cmark & \cmark & \xmark & \xmark & \xmark & \xmark & \xmark \\
 \citet{omar} & \cmark & \xmark & \xmark & \xmark & \xmark & \xmark & \xmark & \cmark & \cmark & \cmark & \xmark & \xmark \\
 \citet{madt}  & \cmark & \xmark & \xmark & \xmark & \xmark & \xmark & \xmark & \cmark & \xmark & \xmark & \cmark & \cmark \\
 \citet{cfcql} & \cmark & \cmark & \cmark & \cmark & \xmark & \xmark & \xmark & \xmark & \xmark & \xmark & \xmark & \xmark \\
 \citet{omiga} & \xmark & \xmark & \cmark & \cmark & \xmark & \xmark & \xmark & \xmark & \xmark & \cmark & \xmark & \cmark \\
\bottomrule
\end{tabular}
\end{adjustbox}
\end{table}

\begin{table}
\centering
\scriptsize
\setstretch{1.5}
\caption{Summary of the evaluation methodologies from the five papers in our case study.}\label{tab:perform_report}
\vspace{1em}
\begin{adjustbox}{max width=\textwidth}
\begin{tabular}{c|cccc}
\textbf{Paper} & \textbf{Evaluation frequency} & \textbf{Performance metrics} & \textbf{Results given as}  & \textbf{Seeds} \\

\hline 
\citet{maicq} % done
& $10$ episodes per $50$ training episodes
 & \begin{tabular}{c} 
Episode return
\end{tabular} &
Plots &
$5$
\\

\hline
\cite{omar}
 & \begin{tabular}{c} 
Not reported
\end{tabular} &
\begin{tabular}{c} 
    Normalised score
\end{tabular} &
Plots
& 5
\\

\hline 
\cite{madt} %done
& \begin{tabular}{c} 
$32$ evaluation epochs at points during training    
\end{tabular} 
 & \begin{tabular}{c}
 Episode return 
\end{tabular} & Plots & Not reported \\

\hline
\citet{cfcql}
& Not reported
& \begin{tabular}{c} 
Episode return, normalised score
\end{tabular} & Tabulated Values & 5 \\

% \hline
% OMAC %done
% & 32 evaluation episodes at points during training
% & \begin{tabular}{c} 
% Episode return \\ 
% Sample efficiency
% \end{tabular}
% & Mean, standard deviation - plots
% & 5
% \\

\hline
\citet{omiga}
& $32$ episodes, sourcing method not reported
& \begin{tabular}{c} 
Episode return
\end{tabular} & Tabulated Values & 5 \\

\bottomrule
\end{tabular}
\end{adjustbox}
\end{table}

\paragraph{Inconsistencies in Evaluation Methodology} \label{sec:eval}
Finally, we find several discrepancies in the way authors evaluate and present their results. Consider Table~\ref{tab:perform_report} which provides a summary of the approaches taken. Not only are there no dimensions along which evaluation is consistent, there are also gaps in the reporting of evaluation procedures, which make it difficult to compare results across studies. % Normalisation strategies (\citet{omar} and \citet{cfcql} both use the strategy proposed by (cite d4rl)), though consistent in the formulas given, are performed differently in the code (cite).
Furthermore, compared to the 10 seed standard recommended by~\citet{agarwal2022deep} and adopted by the online MARL community~\citep{gorsane2022standardised}, using 5 seeds is most common in the papers from our case study. Owing to these small sample sizes the statistical validity of results could be questioned. However, often even this approximate measure of statistical uncertainty is ignored, where claims are made only based on which algorithms achieve the highest mean return across tasks. Consult the appendices for a visual comparison of the tabulated data for state-of-the-art (SOTA) claims. For many tasks the increase in reported performance is not statistically significant due to overlapping error bounds.

Notably, papers lack consideration for the effect that the computational training and online tuning budget can have on the outcome of reported experiments. Computational budget is especially important. In the online setting, environment interaction is often the most expensive part of training. However, in the offline setting, the greatest computational cost is likely updating the model parameters, and therefore for newly proposed algorithms this should be of great interest. Yet, hardware and time taken to train are typically not reported in papers. 
As for the online tuning budget, in the single-agent case \citet{kurenkov2022onlineEvalBudget} note that this budget can have a significant impact on which offline RL algorithms are preferred across a variety of domains. We assert that this budget is equally important for the multi-agent case, but is not being controlled for by any paper in our case study. 

\begin{figure}
    \begin{subfigure}[b]{0.37\textwidth}
        \centering
        \includegraphics[width=0.95\linewidth]{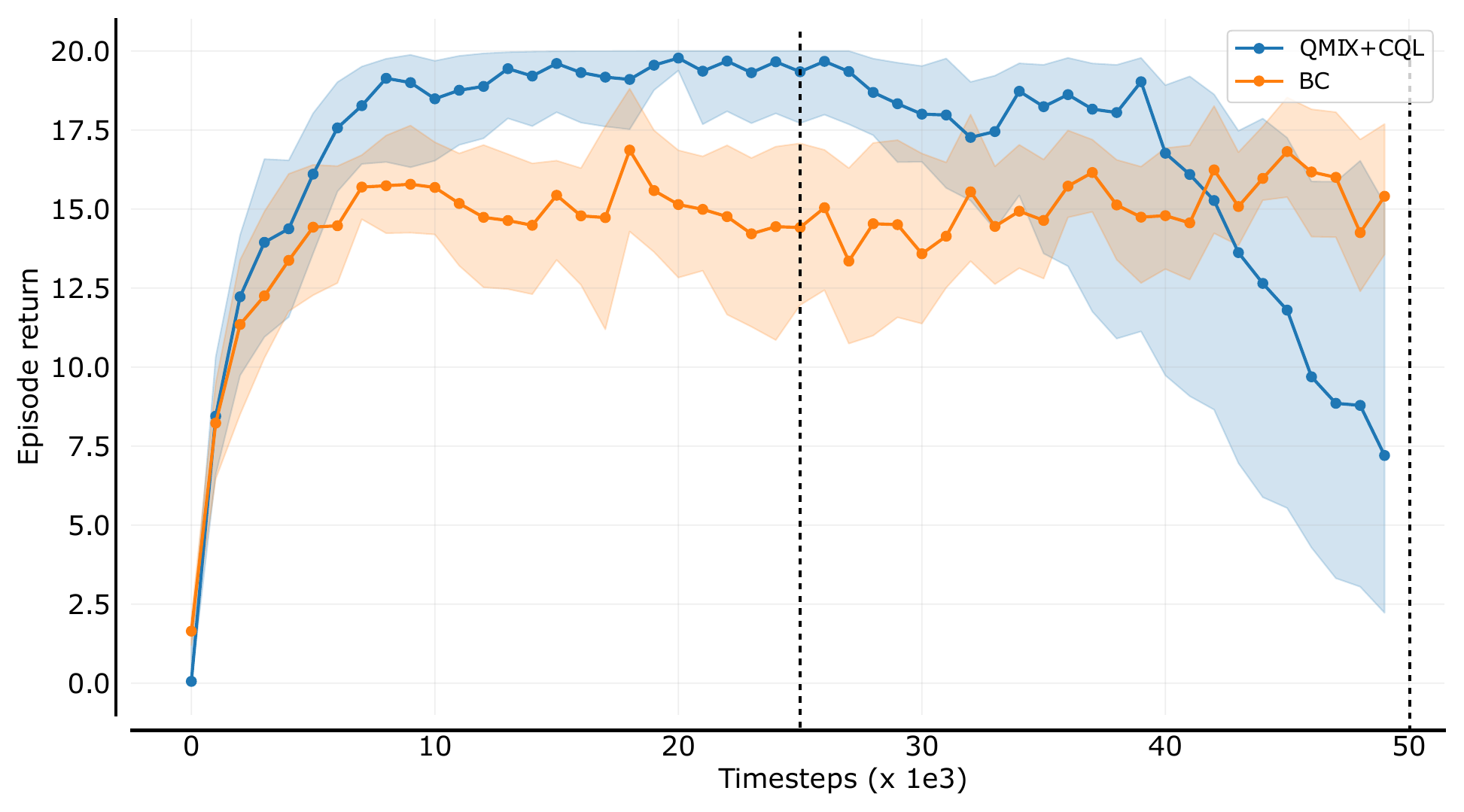}
        \caption{Sample efficiency with bootstrap confidence intervals ~\citep{agarwal2022deep}.}
        \label{fig:sample_efficient_budget}
    \end{subfigure}
    \hfill
    \begin{subfigure}[b]{0.61\linewidth}
        \centering
        \includegraphics[width=\linewidth]{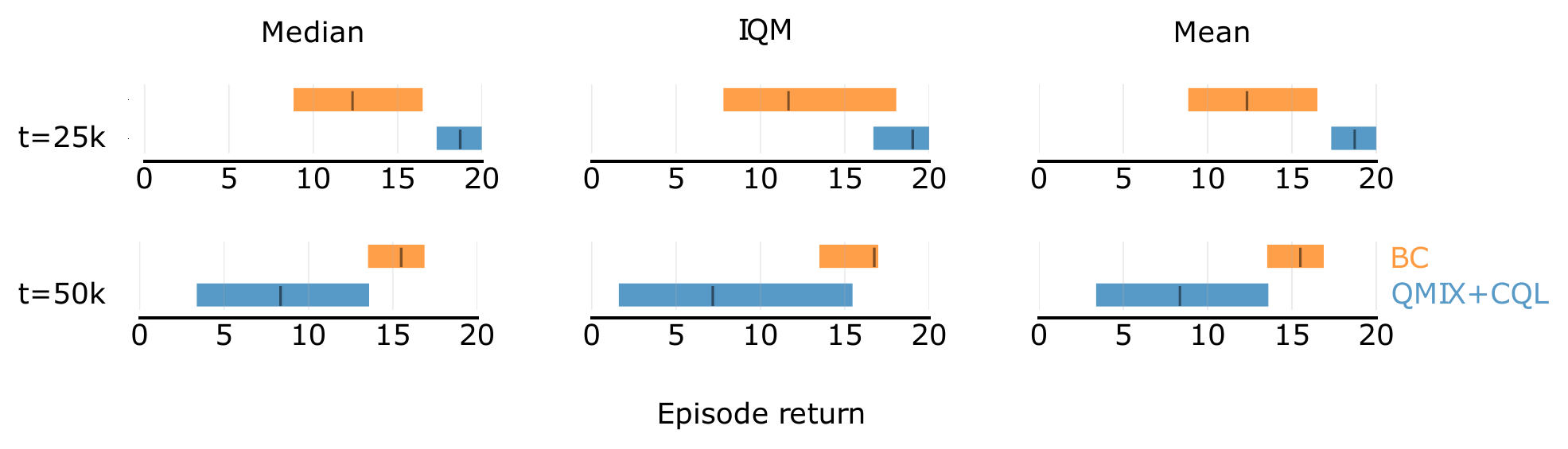}
        \caption{Performance reported at $25k$ steps and at $50k$ steps.}
    \end{subfigure}
    \caption{A comparison of the performance of behaviour cloning (BC) and QMIX+CQL on the SMACv1 \texttt{8m} scenario with the \texttt{Medium} dataset, across 10 seeds. Although QMIX+CQL outperforms BC during the first half of training, its performance deteriorates in the second half, making BC the preferred algorithm over the maximum training time. }
    \label{fig:budget_matters}
\end{figure}

To emphasise the importance of these two budgets, consider Figure~\ref{fig:budget_matters}, where we train two algorithms on the \texttt{8m} SMACv1 scenario. Notice how stopping the training at $25k$ steps yield very different conclusions to stopping at $50k$ steps. The training budget significantly affects the preference between algorithms. However, this can be accounted for using regular online evaluations - for example, if we would like QMIX+CQL to be preferred, we could stop training at $25k$ steps. For online MARL, this is acceptable since access to the environment has no limitations. For the use cases of offline MARL, however, we do not necessarily have access to regular online evaluations. So although we can observe a preferable stopping point, a fair evaluation of an offline algorithm would not be able to use earlystopping unless a large enough evaluation budget is specified. If an online tuning budget is specified and training budget is considered an important hyperparameter, researchers can avoid unintentionally cherry-picking results.

%If we have no online tuning budget, then the recommendation of which algorithm to use does and should change for different options of the `training steps' hyperparameter. We have therefore shown that algorithm superiority is conditional on the online tuning budget and since the hyperparameter we demonstrate this on is training steps, we have also shown that algorithm superiority is conditional on the training budget. %~\citet{madt} consider an online finetuning allowance, but not a hyperparameter tuning allowance for the offline part of their approach.~\citet{omar} use previously finetuned models as baselines and tune specific hyperparameters for two of their algorithms.~\citet{cfcql} only addresses the tuning of one hyperparameter $\tau$ for their proposed algorithm, CFCQL.~\citet{omiga} introduces a hyperparameter $\alpha$ and do not discuss the budget used to tune it.~\citet{maicq} does not mention tuning.

The lack of consistency, transparency, and completeness of evaluation procedures in offline MARL slows progress by forcing researchers to perform expensive re-evaluations of baselines for their comparisons. But perhaps most damaging is the inability to compare and build upon prior work. This allows the field to maintain a mirage of steady progress, while in reality, algorithms are not becoming materially better.
\section{Reevaluating the Empirical Evidence from Prior Work} \label{sec:benchmarking}

Given the problems of baseline and evaluation methodology in offline MARL, we now revisit the empirical evidence from prior work through an independent standardised study. 

Arguably, the gold standard for such a study would use the exact datasets from the authors and their original code (note, even this approach may have drawbacks such as potentially perpetuating poor experimental design and algorithm implementations). Alternatively, we could obtain a selection of similar yet distinct datasets (e.g.~those from~\citet{formanek2023og-marl}), and use existing code implementations from the respective authors for each algorithm considered. Unfortunately, this approach often proves to be infeasible. In some works, only parts of the proposed algorithm code are shared (e.g.~\citet{omar} only share code for their method in continuous action spaces), and in many works, code for the baseline algorithms is omitted completely. As another alternative, we could decide to use our \emph{own} algorithm implementations and datasets, however, this would put us in similar territory as prior work with regards to drawing concrete conclusions. As the most sensible middle ground, we do the following: we use the exact same datasets as provided in prior work, but train our own \emph{baseline} implementations on these datasets; for the results of the other algorithms, we extract the values exactly as they are given in their respective papers. As an illustrative example, suppose we are comparing our implementation of MADDPG with CQL against the results from the OMIGA paper~\citep{omiga}, in one of the scenarios from MAMuJoCo~\citep{peng2021facmac}. Here, we take the \emph{datasets} provided by \citet{omiga}, train our MADDPG+CQL algorithm, and compare these results to the tabular values reported by \citet{omiga} themselves. We feel this methodology is the fairest to the original authors, especially in the situation where the publicly available code is lacking and/or broken. We also view the task of implementing our own baselines, instead of attempting to re-implement the author's proposed algorithm, as a more faithful exercise.

Nonetheless, this approach still has challenges, for it requires access to the datasets used by other authors. Regrettably, in some cases we could not access this data, either because it was never made publicly available, or because the provided download links were broken and multiple attempts to reach the original authors were unsuccessful. In Table~\ref{tab:dataset_access}, we summarise our dataset access record, across the papers in the case study.

\begin{table}
\centering
\scriptsize
\setstretch{1.5}
\caption{Summary of dataset accessibility and whether we benchmarked our baselines on them.}\label{tab:dataset_access}
\vspace{1em}
\begin{adjustbox}{max width=\textwidth}
\begin{tabular}{l|lclc}
\textbf{Paper}& \textbf{Environment} & \textbf{Number of Datasets} & \textbf{Accessibility} &\textbf{Benchmarked} \\ \hline
\cite{maicq} & SMACv1 & 4& {\textcolor{xmarkcolour}{Link broken}}& \xmark \\ \hline
& SMACv1 & 4 & {\textcolor{xmarkcolour}{Not available}} & \xmark\\
& MAMuJoCo & 4 & Yes& \cmark\\
\multirow{-3}{*}{\cite{omar}}& MPE& 12 & Yes& \cmark\\ \hline
\cite{madt}& SMACv1 & 62& {\textcolor{xmarkcolour}{Download fails}}& \xmark\\ \hline
& SMACv1 & 16& Yes & \cmark\\
& MAMuJoCo & 4 & Yes, from \cite{omar} & \cmark \\
\multirow{-3}{*}{\cite{cfcql}} & MPE& 12 & Yes, from \cite{omar} & \cmark \\ \hline
& SMACv1 & 12& Yes& \cmark\\
\multirow{-2}{*}{\cite{omiga}} & MAMuJoCo & 12& Yes& \cmark\\ \hline
\end{tabular}
\end{adjustbox}
\end{table}

The next challenge is that several works use modified versions of environments to generate their datasets, and to evaluate their algorithms. For example, \citet{omiga} modify the MAMuJoCo environment such that agents all receive a global observation of the environment, rather than decentralised, partial observations as is standard. They also use a different version of SMACv1, seemingly first modified by \citet{mappo}, that has important differences from the original. Similarly, \citet{omar} include their own code for the Multi Particle Environments (MPE), which differs from the standardised and maintained version in PettingZoo~\citep{terry2021pettingzoo}. As a consequence, several datasets from different papers \emph{seem} to share a common environment, but in reality do not (e.g., the \texttt{5m\_vs\_6m} datasets generated by \citet{omiga} are not compatible with the \texttt{5m\_vs\_6m} datasets from \citet{cfcql}). To facilitate fair comparisons to each of the original works, we re-use the respective unique environment configuration, even when this is different from the standard. \emph{We do not advocate this approach for future work} and note that standardisation is crucial going forward (which we discuss in more detail in the next section).

We use four baselines for our experiments---two for discrete action spaces, and two for continuous. In the discrete case, we implement BC, and independent Q-learners (IQL)~\citep{tampuu2017multiagent} with CQL regularisation to stabilise offline training. Notably, these are very straightforward baselines that do not rely on any value factorisation or global state information. In continuous action spaces, we use independent DDPG agents with behaviour cloning regularisation---a naive multi-agent extension of the algorithm by \citet{fujimoto2021minimalist}, which notably only requires a single line to change in our vanilla implementation of independent DDPG.
Finally, we also test MADDPG~\citep{lowe2017maddpg} with CQL to stabilise offline critic learning. Interestingly, this baseline is used in multiple works~\citep{omar, omiga}, but its reported performance is poor.

We train our baselines on MPE, SMACv1, and MAMuJoCo scenarios for $50k$, $100k$, and $200k$ training updates, respectively. At the end of training, we compute the mean episode return over 32 episodes and repeat each run across 10 independent random seeds. We avoid fine-tuning our algorithms on each scenario independently in an attempt to control for the online tuning budget~\citep{kurenkov2022onlineEvalBudget}, and instead keep the hyperparameters fixed across each respective scenario.

\paragraph{Result} We provide all the experimental results in tabular form in the appendix. From our own training results, we provide the mean episode return with the standard deviation across 10 independent seeds. As stated above, we extract values verbatim from prior work for other algorithms that are being compared. We perform a simple heteroscedastic, two-sided t-test with a 95\% confidence interval for testing statistical significance, following~\citet{papoudakisBenchmarkingMultiAgentDeep2021}. We summarise the tabulated results in an illustrative plot in Figure \ref{fig:comp_results} (plotting our best baseline per dataset). To standardise comparisons, we normalise results from each dataset (i.e.~scenario-quality-source combination) by the SOTA performance from the literature for that dataset. When our method is significantly better or equal to the best method in the literature, we indicate so using a star. We find that on $35$ out of the $47$ datasets tested, we match or surpass the performance of the current SOTA from the literature.\footnote{Code used to process results is available in a notebook: \href{https://tinyurl.com/offline-marl-meta-review}{https://tinyurl.com/offline-marl-meta-review}}
\section{Standardising Baselines and Evaluation}

The outcome from our benchmarking exercise paints a worrying picture of the state of offline MARL. We maintain that most of the contributions made by the research community to date are valuable. However, because several works seem to be building upon unreliable baselines and using inconsistent evaluation protocols, the overall value to the community is diminished. We believe the community will be better served if we adopt a common set of datasets, baselines, and evaluation methodologies.

% This reinforces our claim in Section~\ref{sec:method_problems} that the application of non-standardised baselines and inconsistent evaluation protocols are making it challenging to accurately measure progress in the field. 

\paragraph{Datasets} With regards to common datasets, OG-MARL \citep{formanek2023og-marl} includes a wide range of offline MARL datasets which the community has begun to adopt \citep{zhu2023madiff, yuan2023survey}. We find that a notable advantage of OG-MARL datasets, aside from their ease of accessibility, is that they are generated on the standard environment configurations rather than customised ones. This significantly eases the challenge of matching datasets to environments, as highlighted in Section~\ref{sec:benchmarking}. Having said that, we also believe there is value in improving access to the datasets from prior works, which we used here for benchmarking. Thus, we convert all of the datasets available to us from the literature (see Table~\ref{tab:dataset_access}) to the OG-MARL dataset API to make them more easily available to the community, in one place. We include statistical profiles of the datasets in the appendix and give credit to the original creators. 

\paragraph{Baselines} While notable progress has been made standardising offline MARL datasets, we maintain that inconsistent use of baselines remains an overlooked issue in the field. Indeed, to highlight this, we conducted extensive benchmarking in Section~\ref{sec:benchmarking}. Now, to address the issue, we release our implementations of BC, IQL+CQL, IDDPG+BC, and MADDPG+CQL, as high-quality baselines for future research. Our baselines come with three main advantages. First, their demonstrated correctness, where we have shown they perform as well as, or better than, most algorithms in the literature on a wide range of datasets. Second, our baselines are easy to parse while also being highly performant on hardware accelerators. The core algorithm logic of our baselines is contained within a single file, which makes it easy for researchers to read, understand, and modify. In addition, all of the training code can be \texttt{jit} (just-in-time) compiled to XLA, making it very fast on hardware accelerators. Furthermore, we use the hardware accelerated replay buffers from Flashbax~\citep{flashbax}, delivering performance gains by speeding up the time to sample from datasets. The third advantage is their compatibility with OG-MARL datasets, which comes ``out of the box'', offering the widest compatibility with offline MARL datasets in the literature~\footnote{Datasets and baselines can be accessed at  \url{https://github.com/instadeepai/og-marl}}. As a foundation for future work, we provide tabular performance values across multiple scenarios for these algorithms, in Table~\ref{tab:new-foundations-results}. Furthermore, all raw training results can be viewed and downloaded, and are linked to in the appendix.

\begin{table}
\setstretch{1.35}
\centering
\caption{Three return metrics---the Final, $\langle$Maximum$\rangle$, (Average)---from the two baseline algorithms in two respective environments, shown across various scenario and dataset quality combinations. Each result is presented as the mean and standard deviation, over 10 seeds. For the Final return, boldface indicates the best performing algorithm, and an asterisk (*) indicates a metric is not significantly different from the best performing metric in that situation, based on a heteroscedastic, two-sided t-test with 5\% significance.
}
\vspace{1em}
\label{tab:new-foundations-results}
\begin{subtable}{0.411\linewidth}
\caption{SMACv1}
\adjustbox{width=\textwidth}{
\begin{tabular}{cc cc}
\textbf{Scenario} & \textbf{Quality} & \textbf{BC} & \textbf{IQL+CQL} \\ \hline
%%%%%%%%%%%%%%%%%%%%%%%%%%%%%%%%%%%%%%%%%%
& Good &
\RESULT{14.94}{1.58}{19.60}{0.90}{16.18}{3.19} &
\RESULT{\textbf{20.00}}{\textbf{0.00}}{20.00}{0.00}{18.80}{2.79}
\\ \cline{2-4}
%%%%%%%%%%%%%%%%%%%%%%%%%%%%%%%%%%%%%%%%%%%%
8m & Medium &
\RESULT{10.65}{1.41}{11.32}{1.77}{9.71}{1.93} &
\RESULT{\textbf{19.10}}{\textbf{1.24}}{20.00}{0.00}{18.94}{2.82}
\\ \cline{2-4}
%%%%%%%%%%%%%%%%%%%%%%%%%%%%%%%%%%%%%%%%%%%%
& Poor &
\RESULT{\textbf{5.35}}{\textbf{0.44}}{5.59}{0.20}{5.20}{0.55} &
\RESULT{4.85}{0.12^*}{5.06}{0.31}{4.81}{0.48}
\\ \hline
%%%%%%%%%%%%%%%%%%%%%%%%%%%%%%%%%%%%%%%%%%
& Good &
\RESULT{18.18}{1.06^*}{19.44}{1.30}{17.44}{2.53} &
\RESULT{\textbf{19.60}}{\textbf{0.92}}{20.03}{0.05}{19.17}{2.64}
\\ \cline{2-4}
%%%%%%%%%%%%%%%%%%%%%%%%%%%%%%%%%%%%%%%%%%%%
2s3z & Medium &
\RESULT{13.14}{2.03}{13.99}{2.78}{12.04}{1.91} &
\RESULT{\textbf{15.79}}{\textbf{0.98}}{18.04}{1.38}{16.00}{2.53}
\\ \cline{2-4}
%%%%%%%%%%%%%%%%%%%%%%%%%%%%%%%%%%%%%%%%%%%%
& Poor &
\RESULT{6.42}{1.39^*}{8.16}{0.53}{6.57}{1.18} &
\RESULT{\textbf{7.85}}{\textbf{1.19}}{9.47}{0.64}{8.50}{1.33}
\\ \hline
%%%%%%%%%%%%%%%%%%%%%%%%%%%%%%%%%%%%%%%%%%
& Good &
\RESULT{15.67}{3.49^*}{17.88}{1.19}{15.31}{3.35} &
\RESULT{\textbf{16.19}}{\textbf{1.64}}{17.87}{2.17}{14.33}{3.88}
\\ \cline{2-4}
%%%%%%%%%%%%%%%%%%%%%%%%%%%%%%%%%%%%%%%%%%%%
5m\_vs\_6m & Medium &
\RESULT{10.92}{0.93}{14.22}{2.00}{11.52}{2.73} &
\RESULT{\textbf{15.78}}{\textbf{2.93}}{17.65}{1.76}{13.22}{3.46}
\\ \cline{2-4}
%%%%%%%%%%%%%%%%%%%%%%%%%%%%%%%%%%%%%%%%%%%%
& Poor &
\RESULT{7.33}{0.55}{8.38}{1.51}{7.13}{1.05} &
\RESULT{\textbf{10.87}}{\textbf{2.63}}{12.21}{2.21}{9.99}{2.29}
\\ \bottomrule
%%%%%%%%%%%%%%%%%%%%%%%%%%%%%%%%%%%%%%%%%%
\end{tabular}
}
\end{subtable}
\hfill
\begin{subtable}{0.53\linewidth}
\caption{MAMuJoCo}
\adjustbox{width=\textwidth}{
\begin{tabular}{cc cc}
\textbf{Scenario} & \textbf{Quality} & \textbf{IDDPG+BC} & \textbf{MADDPG+CQL} \\ \hline
%%%%%%%%%%%%%%%%%%%%%%%%%%%%%%%%%%%%%%%%%%
& Good &
\RESULT{485.11}{508.76}{965.37}{11.74}{284.76}{378.79} &
\RESULT{\textbf{1803.24}}{\textbf{546.31}}{1776.19}{770.78}{1296.62}{727.39}
\\ \cline{2-4}
%%%%%%%%%%%%%%%%%%%%%%%%%%%%%%%%%%%%%%%%%%%%
2x4 Ant & Medium &
\RESULT{890.66}{234.86^*}{973.34}{23.98}{724.09}{264.01} &
\RESULT{\textbf{1052.75}}{\textbf{182.30}}{1067.96}{129.09}{922.47}{181.97}
\\ \cline{2-4}
%%%%%%%%%%%%%%%%%%%%%%%%%%%%%%%%%%%%%%%%%%%%
& Poor &
\RESULT{\textbf{888.83}}{\textbf{110.97}}{992.16}{54.86}{891.38}{108.36} &
\RESULT{504.24}{125.56}{979.42}{10.95}{479.40}{157.92}
\\ \hline
%%%%%%%%%%%%%%%%%%%%%%%%%%%%%%%%%%%%%%%%%%
& Good &
\RESULT{\textbf{266.57}}{\textbf{75.26}}{978.47}{2.93}{401.17}{282.74} &
\RESULT{103.36}{604.36^*}{1191.14}{641.59}{540.98}{628.38}
\\ \cline{2-4}
%%%%%%%%%%%%%%%%%%%%%%%%%%%%%%%%%%%%%%%%%%%%
4x2 Ant & Medium &
\RESULT{969.96}{175.01^*}{1318.59}{194.35}{1109.58}{213.13} &
\RESULT{\textbf{1394.98}}{\textbf{350.67}}{1477.25}{141.49}{1302.99}{266.38}
\\ \cline{2-4}
%%%%%%%%%%%%%%%%%%%%%%%%%%%%%%%%%%%%%%%%%%%%
& Poor &
\RESULT{\textbf{838.84}}{\textbf{59.49}}{975.24}{10.54}{861.92}{85.00} &
\RESULT{-1463.67}{1627.83}{966.86}{8.75}{-841.30}{1322.39}
\\ \hline
%%%%%%%%%%%%%%%%%%%%%%%%%%%%%%%%%%%%%%%%%%
& Good &
\RESULT{\textbf{5411.68}}{\textbf{501.33}}{5733.50}{290.89}{4765.36}{1433.66} &
\RESULT{3876.05}{995.80}{2988.31}{1658.94}{1735.93}{1811.54}
\\ \cline{2-4}
%%%%%%%%%%%%%%%%%%%%%%%%%%%%%%%%%%%%%%%%%%%%
2x3 HalfCheetah & Medium &
\RESULT{\textbf{2819.67}}{\textbf{106.05}}{2955.68}{240.16}{2482.85}{511.73} &
\RESULT{2330.55}{277.40}{2489.85}{321.24}{2204.64}{451.13}
\\ \cline{2-4}
%%%%%%%%%%%%%%%%%%%%%%%%%%%%%%%%%%%%%%%%%%%%
& Poor &
\RESULT{\textbf{676.65}}{\textbf{59.43}}{703.75}{26.57}{630.68}{107.89} &
\RESULT{-35.92}{32.53}{-4.72}{4.15}{-46.00}{77.23}
\\ \bottomrule
%%%%%%%%%%%%%%%%%%%%%%%%%%%%%%%%%%%%%%%%%%
\end{tabular}
}
\end{subtable}
\end{table}

\paragraph{Evaluation}
% A lack of standardised evaluation protocols has been a challenge in both online and offline MARL. 
Recent efforts have been made to address the issue in the online setting \citep{gorsane2022standardised, agarwal2022deep}. However, similar efforts are still absent in the offline setting, as discussed in Section~\ref{sec:eval}. Following in the spirit of this work, we propose a set of evaluation guidelines for offline MARL, given in the blue box on Page~\pageref{box:eval}, which we believe will significantly improve research outcomes, if adopted.

\begin{figure}
    \centering
    \begin{subfigure}[t]{0.45\textwidth}
        \includegraphics[width=\textwidth]{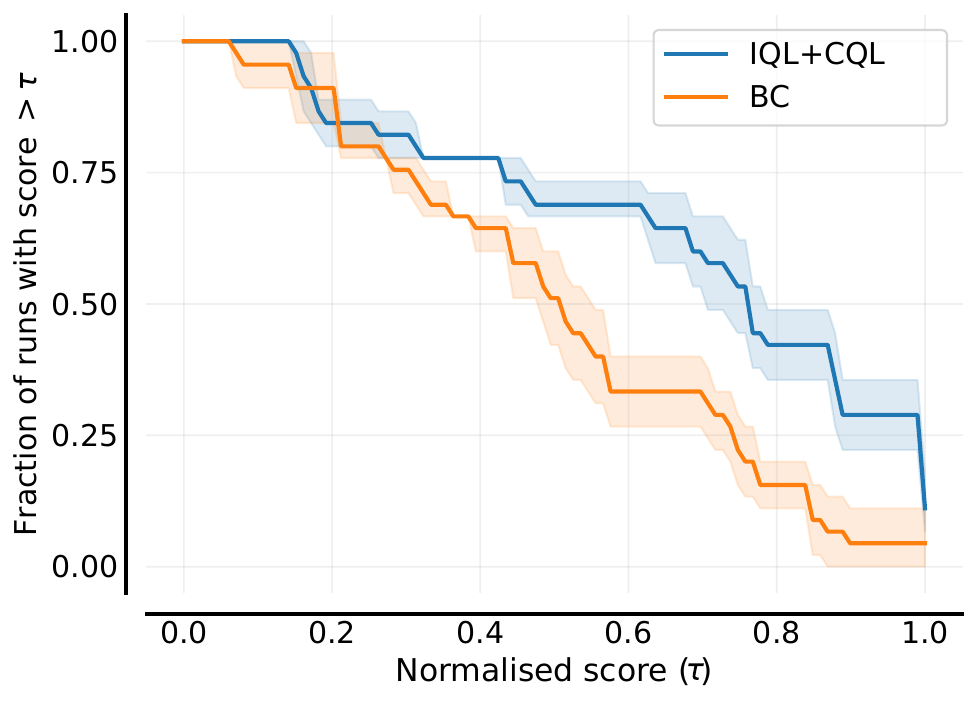}
        \caption{SMACv1}
        \label{fig:enter-label}
    \end{subfigure}
    \hspace{1cm}
    \begin{subfigure}[t]{0.45\textwidth}
        \includegraphics[width=\textwidth]{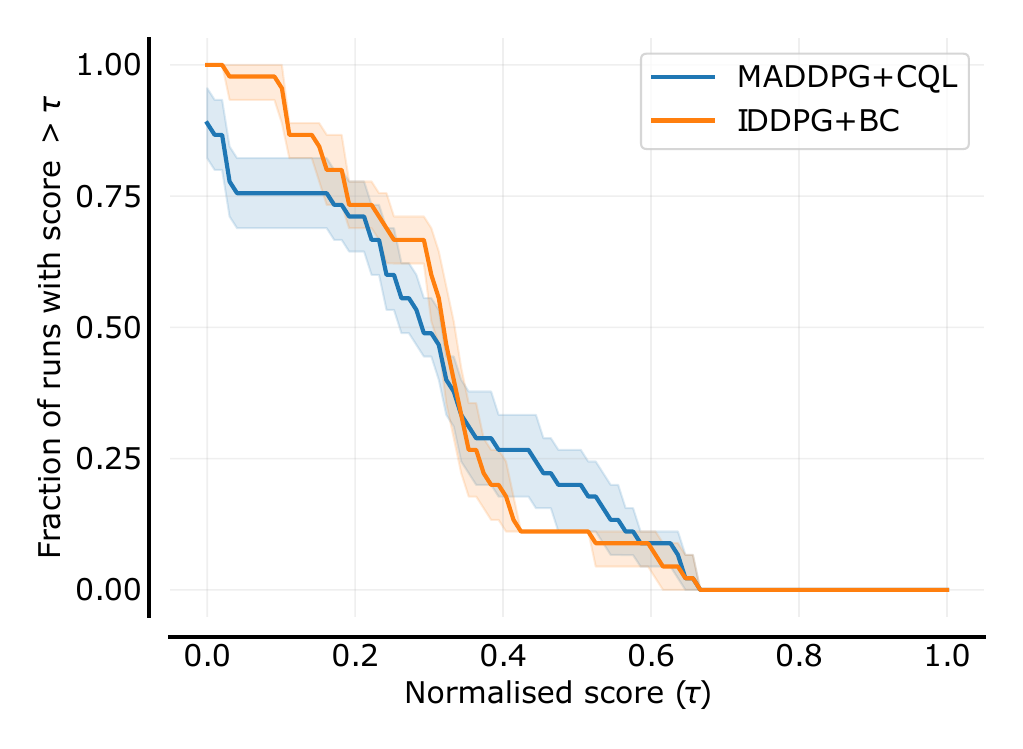}
        \caption{MAMuJoCo}
        \label{fig:enter-label}
    \end{subfigure}
    \caption{Performance profiles \citep{agarwal2022deep} aggregated across all results from Table~\ref{tab:new-foundations-results} on SMACv1 and MAMuJoCo. Scores are normalised as per~\citet{fu2021d4rl}.}
    \label{fig:performance_profiles}
\end{figure}

\begin{mybox}{\small Evaluation Guidelines for Cooperative Offline MARL}{box:eval}
\small
\textbf{Choosing the settings to evaluate on:}
\begin{itemize}[leftmargin=10pt]
    \item Select at least 2-3 different environments on which to test. For example, evaluating on both SMAC and MAMuJoCo is the most common combination. We encourage additionally evaluating on environments beyond these two, to avoid overfitting to them. Do not use non-standard environment configurations without explicitly stating so.
    \item For each environment choose at least 3-4 different scenarios. If the environment creators provide a minimal set of recommended scenarios, use those. Alternatively, focus on scenarios that are common in prior literature.
    \item Choose a range of dataset quality types. We recommend at least including a "good" dataset where the majority of samples are from good policies and a "mixed" dataset where samples come from a wide range of policies including good, medium and poor ones.
    \item Try to use common existing datasets from the literature \citep{formanek2023og-marl}. If you include your own dataset, provide a clear reason for why, and make it easily accessible to the community.
\end{itemize}

\textbf{Choosing the baselines to compare to:}
\begin{itemize}[leftmargin=10pt]
    \item Choose at least 3-4 relevant baselines to compare to. 
    \item Usually include behaviour cloning, especially on good datasets where it can be challenging to beat. 
    \item Try to use common and existing implementations of baselines from the literature.
    \item If you include your own novel baseline, make the code available and easy to run for future comparisons. It is not sufficient to only share the code for the novel algorithm being proposed.
\end{itemize}

\textbf{Choosing the training and evaluation parameters:}
\begin{itemize}[leftmargin=10pt]
    \item For each environment, set a training budget and keep it constant across algorithms. For example, we used 100k training updates on SMAC and 200k on MAMuJoCo.
    \item If possible, do regular evaluations during training so that you can plot a training curve for analysis at the end of training. Do not use information from these evaluations to influence a training run online, as this would violate the assumptions around the training being offline.
    \item We recommend at every evaluation step unrolling for 32 episodes and reporting the average episode return.
    \item As per \cite{agarwal2022deep}, you should repeat each run across 10 random seeds.
\end{itemize}

\textbf{Reporting your results:}
\begin{itemize}[leftmargin=10pt]
    \item Report the final evaluation result, averaged across all 10 seeds, along with the standard deviation.
    \item If doing regular evaluations during training, also report the average and maximum episode return during training as per \cite{papoudakisBenchmarkingMultiAgentDeep2021} and plot sample efficiency curves as per \cite{gorsane2022standardised}.
    \item Use appropriate statistical significance calculations to report on whether your algorithms significantly outperform the baselines \citep{papoudakisBenchmarkingMultiAgentDeep2021, agarwal2022deep}.
    \item In addition to reporting results on a per-dataset basis, also report aggregated results across scenarios and dataset types. Results can be aggregated by first normalising them as per \cite{fu2021d4rl} and then applying the utilities from \textit{MARL-eval} \citep{gorsane2022standardised} (e.g. performance profile plots, see Figure \ref{fig:performance_profiles}).  
\end{itemize}
\end{mybox}

\section{Conclusion}
We conducted a thorough analysis of prior work in offline MARL and identified several significant methodological failures which we demonstrated are inhibiting progress in the field. Furthermore, we benchmarked simple baselines against several proposed SOTA algorithms and showed that our baselines outperform them in most cases. We used these insights to propose improving standards in evaluation with a simple protocol.
% I think we should speak about the road ahead -- the reason this is significant = now we can make real progress, building on strong foundations. The goal is not to stop at baselines

\paragraph{Limitations} Our work highlights some important challenges faced by the field of offline MARL, but does not capture \emph{all} such challenges. We hope our contributions will make it easier for authors to align and compare their future work, but we ultimately realise that our efforts require vetting by the community, to be tested and validated over time. We welcome such engagements, to collectively chart a path forward.

% \paragraph{Limitations}
% TODO 

% \paragraph{Future work}
% TODO

\newpage
\bibliographystyle{abbrvnat}
\bibliography{references}

\newpage
\appendix

\section{Algorithm implementation details and hyperparameters}

\begin{mybox}{\small Machine Learning Reproducibility Checklist (Algorithms, Code, and Experiment Details)}{box:ml_checklist}
\small
\begin{enumerate}
\item For all models and algorithms presented, check if you include:
\begin{enumerate}
  \item A clear description of the mathematical setting, algorithm, and/or model.
    \answerYes{}
  \item A clear explanation of any assumptions.
    \answerYes{}
  \item An analysis of the complexity (time, space, sample size) of any algorithm.
    \answerNo{The algorithms presented are foundational baselines that draw on existing works from the literature.}
\end{enumerate}

\item For all shared code related to this work, check if you include:
\begin{enumerate}
  \item Specification of dependencies.
    \answerYes{We provide a \textit{requirements.txt} file and detailed installation instructions. In addition, we provide a working Dockerfile for maximum portability.}
  \item Training code.
    \answerYes{All systems can be run using a single script. Instructions are provided in the README.}
  \item Evaluation code.
    \answerYes{All systems have inbuilt evaluation and results are logged to the terminal and \textit{Weights and Biases}.}
  \item (Pre-)trained model(s).
    \answerNo{Pre-trained models have not been saved because training a model on any of the scenarios takes between a few minutes and at most a couple hours on a Laptop GPU (RTX 3070).}
  \item README file includes table of results accompanied by precise command to run to produce
those results.
    \answerYes{We provide a notebook with a database of results and visualisations and easy-to-run code for reproducing all results.}
\end{enumerate}

\item For all reported experimental results, check if you include:
\begin{enumerate}
  \item The range of hyper-parameters considered, method to select the best hyper-parameter
configuration, and specification of all hyper-parameters used to generate results.
    \answerYes{See below.}
  \item The exact number of training and evaluation runs.
    \answerYes{Each experiment was repeated across 10 random seeds. For evaluation we rolled out policies for 32 episodes and computed the mean episode return.}
  \item A clear definition of the specific measure or statistics used to report results.
    \answerYes{We measured episode return in all cases.}
  \item A description of results with central tendency (e.g. mean) \& variation (e.g. error bars).
    \answerYes{}
  \item The average runtime for each result, or estimated energy cost.
    \answerYes{Depending on the scenario, reproducing a single experimental run can take between 20min and 4 hours on a Laptop GPU (e.g. RTX 3070). The MPE experiments are the fastest, followed by SMAC experiments and finally MAMuJoCo experiments take the longest.}
  \item A description of the computing infrastructure used.
    \answerYes{Individual runs can easily be reproduced on a Laptop GPU (e.g. RTX 3070), 8GB of RAM and 4 CPU cores. However, reproducing all experiments on a single GPU would take approximately 20 days. We had access to a compute cluster with 10, roughly equivalent, GPUs. This meant that generating all baseline results took about 2 days.}
\end{enumerate}

\end{enumerate}
\end{mybox}

\subsection{IDDPG+BC (Continuous)}
Our implementation draws on the minimalistic offline RL algorithm proposed by \cite{fujimoto2021minimalist}. In essence, you simply add a behaviour cloning term to the deterministic policy gradient (DPG) loss. In a continuous action space with deterministic policies, this can be achieved by a simple mean square error between the output of the policy and the given action sampled from the dataset. As per \cite{fujimoto2021minimalist}, we normalise the DPG term so that its scale is similar to the BC term and then use a \textit{behaviour cloning weight} hyperparameter to control the relative importance of the behaviour cloning term vs. the DPG term. The critic conditioned on the environment state and the agents individual action only. Policies conditioned on the decentralised observations only. We used shared parameters by always concatenating an agent-ID to observations.

\begin{table}[h]
\centering
\caption{Hyper parameters used for IDDPG+BC across all MAMuJoCo and MPE datasets. We found that the recommended \textit{behaviour cloning weight} of 2.5 proposed by \cite{fujimoto2021minimalist} worked well across all scenarios.}
\begin{tabular}{l|l}
\textbf{Hyperparameter}    & \textbf{Value} \\ \hline
Critic first linear layer  & 128            \\
Critic second linear layer & 128            \\
Policy linear layer        & 64             \\
Policy GRU layer           & 64             \\
Critic learning rate       & 1e-3           \\
Policy learning rate       & 3e-4           \\
Target update rate         & 0.005          \\
Discount (gamma)           & 0.99           \\
BC weight                  & 2.5            \\
\end{tabular}
\end{table}

\subsection{MADDPG+CQL (Continuous)}
Our implementation adds CQL \citep{kumar2020cql} to MADDPG \citep{lowe2017maddpg}. In the original version of CQL they had stochastic policies (soft actor critic), and so, getting actions \textit{near} the current policy was simply a matter of sampling the stochastic policy. Since we had deterministic policies we applied a small amount of Gaussian noise to our actions. The amount of noise is then controlled by a hyperparameter we called \textit{CQL sigma}. The \textit{CQL weight} parameter controls the relative importance of the CQL loss in the overall critic loss. While the critic condition on joint-actions and the environment state, policies conditioned on the decentralised observations only. We used shared parameters by always concatenating an agent-ID to observations.

\begin{table}[h]
\centering
\caption{Hyper parameters used for MADDPG+CQL across all MAMuJoCo and MPE datasets. We found that MADDPG+CQL was sensitive to the value of \textit{CQL sigma} and the optimal value depended on the MuJoCo scenario. For \textit{Ant} scenarios, 0.1 was the best value, while for \textit{Hopper} and \textit{HalfChetah} scenarios the best values were 0.2 and 0.3 respectively. This dependence on the scenario makes sense since it is well known that CQL has a dependence on the action space of the scenario tested on \cite{kumar2020cql}. We found that tuning the \textit{CQL weight} across scenarios could also slightly improve performance but the value 3 worked relatively well across all scenarios. Future works could explore using automatic CQL weight tuning, and stochastic policies (e.g. soft actor critic) to remove the CQL sigma hyper-parameter.}
\begin{tabular}{l|l}
\textbf{Hyperparameter}    & \textbf{Value} \\ \hline
Critic first linear layer  & 128            \\
Critic second linear layer & 128            \\
Policy linear layer        & 64             \\
Policy GRU layer           & 64             \\
Critic learning rate       & 1e-3           \\
Policy learning rate       & 3e-4           \\
Target update rate         & 0.005          \\
Discount (gamma)           & 0.99           \\
Number of CQL actions      & 10             \\
CQL weight                 & 3              \\
CQL sigma                  & 0.1, 0.2, 0.3 
\end{tabular}
\end{table}

\subsection{IQL+CQL (Discrete)}
Our implementation added CQL \citep{kumar2020cql} to independent Q-Learners. Our independent Q-learners use recurrent Q-networks \citep{hausknecht2017deep}. For the CQL loss, we simply sample a \textit{number of CQL actions} randomly from the joint-action space and "push" their Q-values down, while pushing up the Q-values for joint-actions in the dataset The \textit{CQL weight} hyperparameter controls the relative importance of the CQL term in the Q-Learning loss. The Q-networks condition on the decentralised observations only. We used shared parameters by always concatenating an agent-ID to observations.

\begin{table}[h]
\centering
\caption{Hyper parameters used for IQL+CQL across all SMAC datasets.}
\begin{tabular}{l|l}
\textbf{Hyperparameter}    & \textbf{Value} \\ \hline
First linear layer         & 64            \\
GRU layer                  & 64            \\
Learning rate              & 3e-4           \\
Target period              & 200            \\
Discount (gamma)           & 0.99           \\
Number of CQL actions      & 10             \\
CQL weight                 & 2              \\
\end{tabular}
\end{table}

\subsection{Behaviour cloning (Discrete)}
In our behaviour cloning implementation for discrete action spaces, we train policy networks to match the actions in the dataset using a simple categorical crossentropy loss. We use recurrent policy networks which condition on the decentralised observations only. We used shared parameters by always concatenating an agent-ID to observations.

\begin{table}[h]
\centering
\caption{Hyper parameters used for BC across all SMAC datasets.}
\begin{tabular}{l|l}
\textbf{Hyperparameter}    & \textbf{Value} \\ \hline
First linear layer         & 64            \\
GRU layer                  & 64            \\
Learning rate              & 1e-3           \\
Discount (gamma)           & 0.99           \\
\end{tabular}
\end{table}

\newpage
\section{Meta-review: Visualising statistical significance in the literature}

We processed results reported in tabular form by~\citet{omar},~\citet{omiga},~\citet{OMAC}, and~\citet{cfcql}. As previously stated, we perform a simple heteroscedastic, two-sided t-test with a 95\% confidence interval for testing statistical significance. If we accept the null hypothesis, it can be said that with a 95\% confidence interval  For this section of the appendix, we consider only the results reported, and do not include our own baselines. 

We show that many of the results (which form a large part of the evidence for SOTA claims for most of the considered papers) do not indicate a significant difference between performance of the algorithm with the highest mean and the next-best performing algorithm. Each result in each plot which has a red circle around it is equivalent to SOTA within the table in which it is reported.

\subsection{OMAC}

\begin{wrapfigure}{r}{0.7\textwidth}
    \centering
    \includegraphics[width=0.7\textwidth]{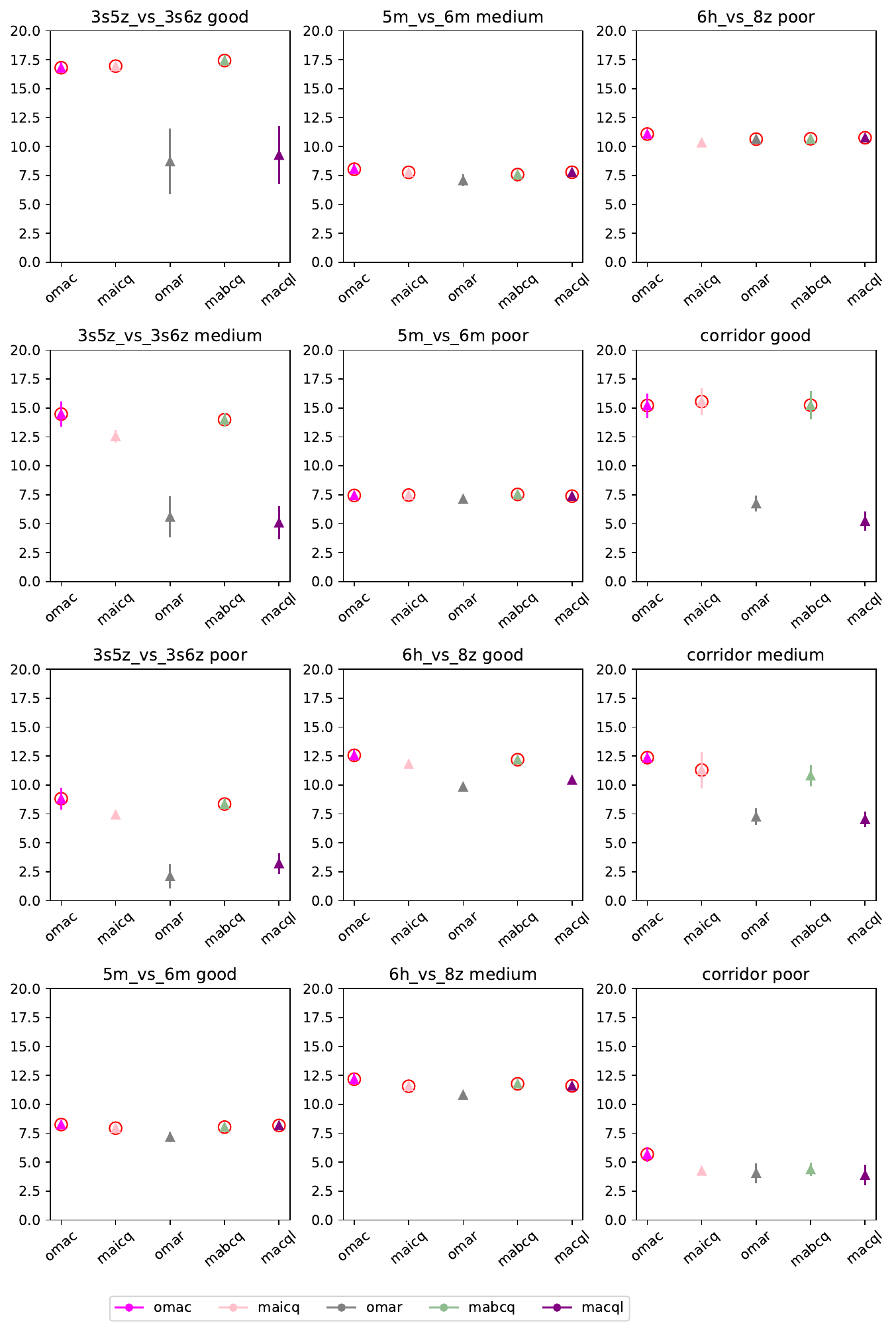}
    \caption{Results reported by ~\citet{OMAC} on the SMAC environment.}
    \label{fig:omac_stats}
\end{wrapfigure}

Figure \ref{fig:omac_stats} illustrates the results presented in Table 4 in the paper by~\citet{OMAC}. We represent SMACv1 results on a $0 - 20$ scale to better interpret results within the scoring range. 
Note the proposed algorithm is unmatched on one dataset only. Additionally, mabcq (not the proposed algorithm) is equivalent to SOTA on all but 2 of the 12 datasets.
\\
\\
\subsection{OMIGA}

\begin{landscape}
\begin{figure}
    \centering
    \includegraphics[width=1.6\textwidth]{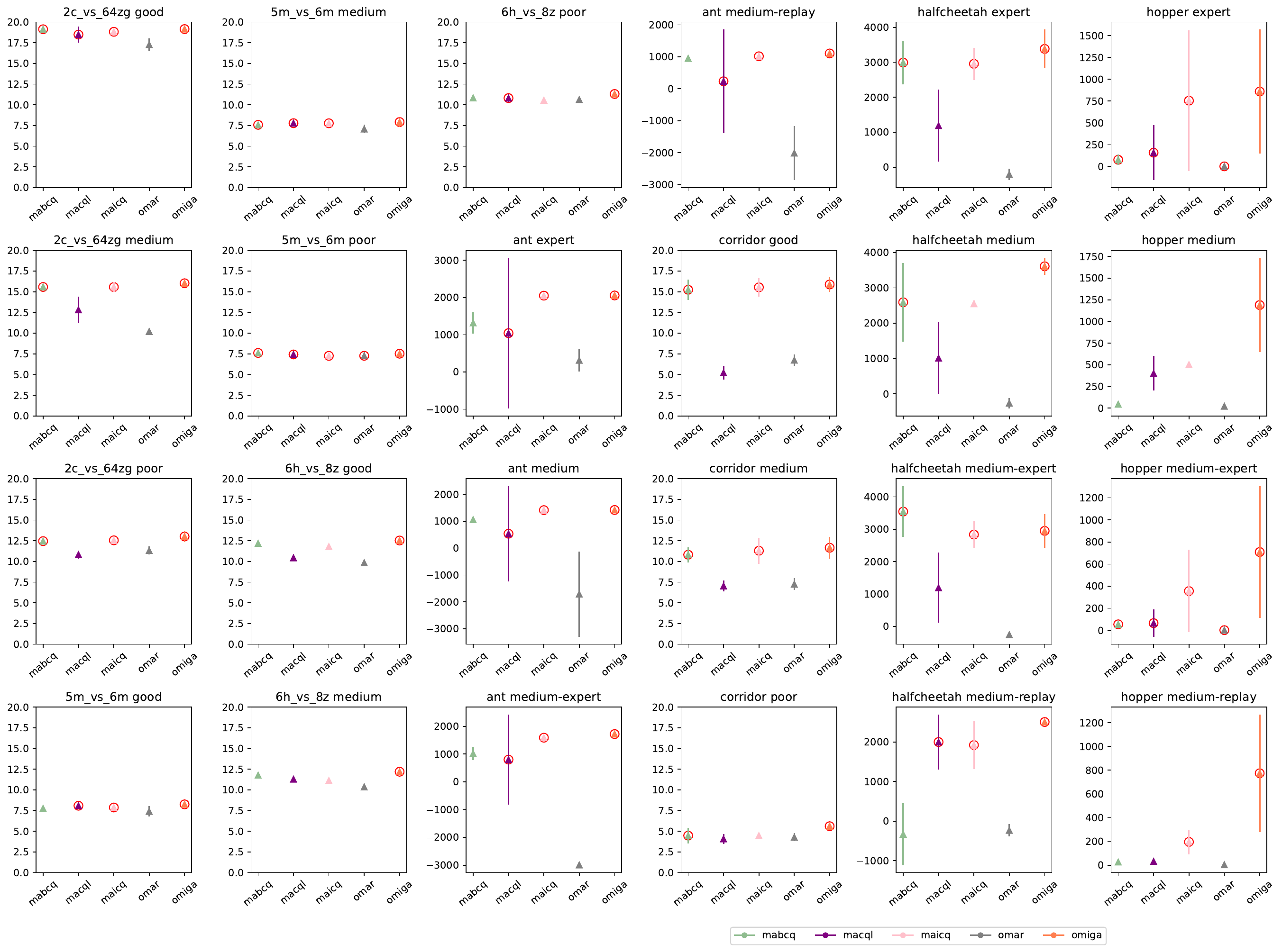}
    \caption{Results reported by~\citet{omiga} on the SMAC and MAMuJoCo environments.}
    \label{fig:omiga_stats}
\end{figure}
\end{landscape}

Figure \ref{fig:omiga_stats} illustrates the results presented in Table 1 in the paper by~\citet{omiga}. We represent SMACv1 results on a $0 - 20$ scale to better interpret results within the scoring range. MAMuJoCo results are unnormalised.

Note the proposed algorithm is unmatched on only 3 of the 24 datasets. Additionally, maicq (not the proposed algorithm) is equivalent to SOTA on 18 of the 24 datasets.

\subsection{OMAR}

\begin{wrapfigure}{r}{0.7\textwidth}
    \centering
    \includegraphics[width=0.7\textwidth]{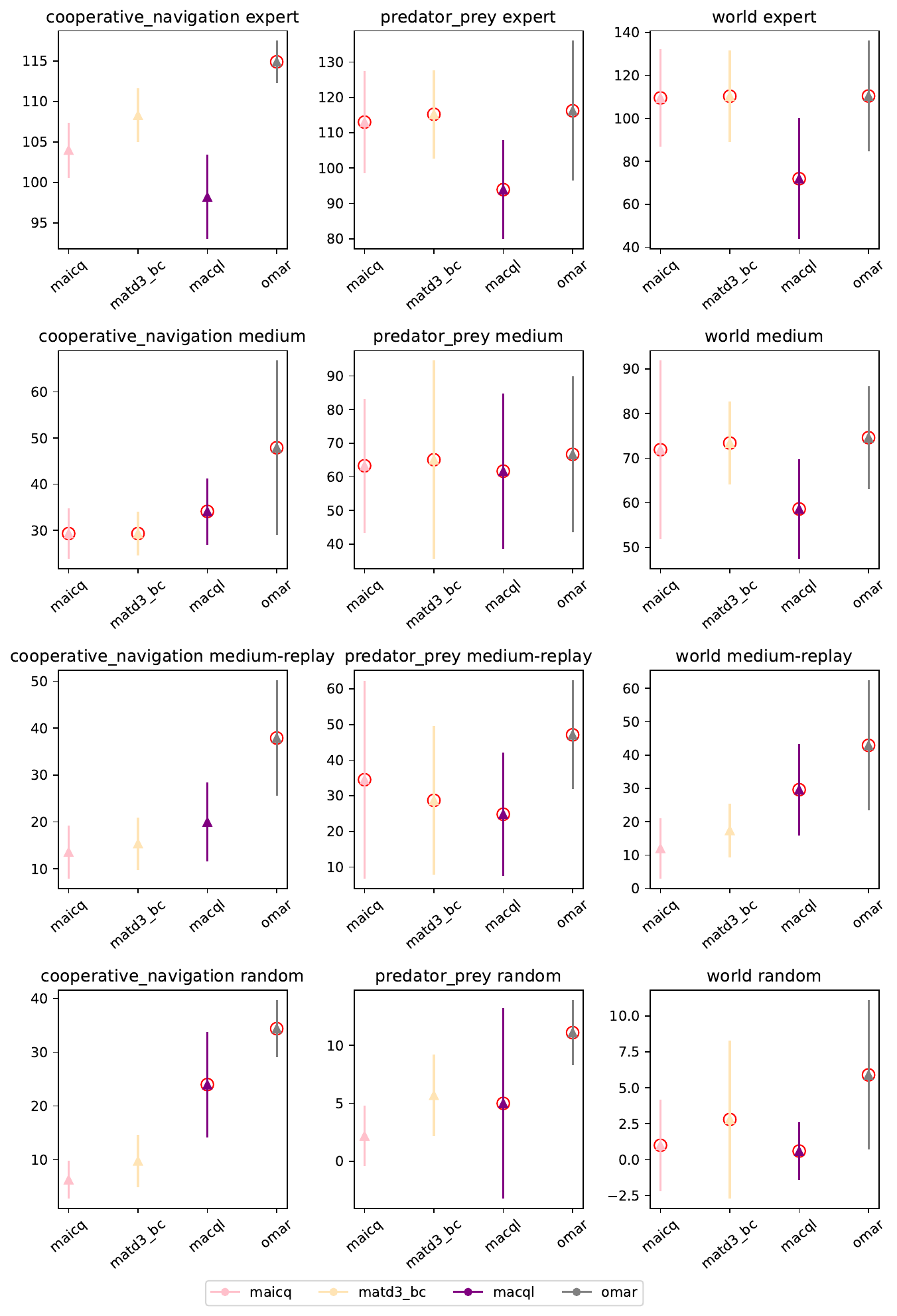}
    \caption{Results reported by~\citet{omar} on the MPE environment.}
    \label{fig:omar_stats}
\end{wrapfigure}

Figure \ref{fig:omar_stats} illustrates the results presented in Table 1 in the paper by~\citet{omar}. MPE results are normalised.

Note the proposed algorithm is unmatched on only 2 of the 12 datasets. Additionally, macql (not the proposed algorithm) is equivalent to SOTA on 10 of the 12 datasets.

\subsection{CFCQL}

Figure \ref{fig:omar_stats} illustrates the results presented in Table 1 in the paper by~\citet{cfcql}. SMAC results are given in terms of win rate. MPE and MAMuJoCo results are normalised.

Note the proposed algorithm is unmatched on only 10 of the 32 datasets.

\begin{landscape}

\begin{figure}
    \centering
    \includegraphics[width=1.6\textwidth]{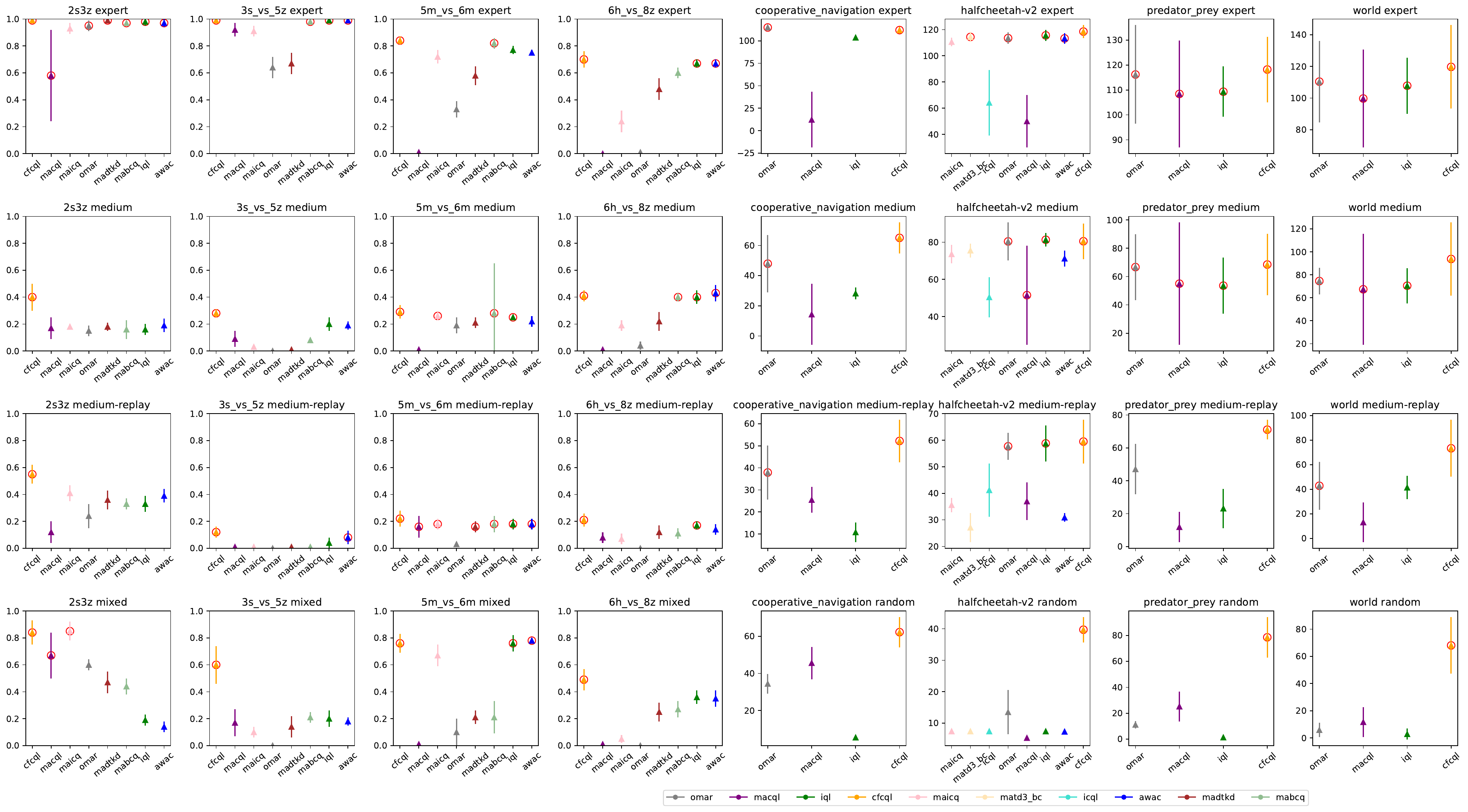}
    \caption{Results reported by ~\citet{cfcql} in tabular form on SMAC, MPE and MAMuJoCo environments.}
    \label{fig:enter-label}
\end{figure}
\end{landscape}

\section{Meta-review: Benchmarking literature results against our own baselines}

The tabular results presented below are used to create Figure 1. Mean and standard deviation are provided - in our case, using 10 seeds, in the case of results from the literature, using 5. A bold result indicates that the mean is highest (one way to define SOTA) and a starred result indicates that the result is not statistically significantly worse or better than the bold result.

\subsection{MPE}

Results in the MPE environment are normalised. The given formula for normalisation is $100 \times (S - S_{random})\div(S_{expert} - S_{random})$. We consider the cooperative navigation (CN), predator-prey (PP) and world (WD) scenario datasets. Both ~\citet{cfcql} and~\citet{omar} provide results on these datasets. We were unfortunately unsuccessful in getting access to the PP and WD datasets. Furthermore, we faced challenges getting the customised environment working because it required loading pre-trained policies from the adversaries and there were limited details provided on how to do that. 

\begin{table}[hbtp]
    \centering
    \caption{Mean and standard deviation results from~\citet{omar},~\citet{cfcql} and our baseline (IDDPG+BC) on MPE environment datasets provided by~\citet{omar}.}
    \begin{adjustbox}{max width=\textwidth} 
\begin{tabular}{ll|llll|lll|l}

 & & \multicolumn{4}{c}{Pan et al.} & \multicolumn{3}{c}{Shao et al.} & Our Baseline \\
 & & maicq & matd3 bc & macql & omar & macql & iql & cfcql & iddpg+bc \\
task & dataset quality &  &  &  &  &  &  &  &  \\
\midrule
\multirow[t]{4}{*}{CN} & expert & 104.00$\pm$3.40 & 108.30$\pm$3.30 & 98.20$\pm$5.20 & \textbf{114.90$\pm$2.60} & 12.20$\pm$31.00 & 103.70$\pm$2.50 &  112.00$\pm$4.00* & 104.90$\pm$3.33 \\
 & medium & 29.30$\pm$5.50 & 29.30$\pm$4.80 & 34.10$\pm$7.20 & 47.90$\pm$18.90 & 14.30$\pm$20.20 & 28.20$\pm$3.90 & 65.00$\pm$10.20 & \textbf{102.02$\pm$10.68} \\
 & medium-replay & 13.60$\pm$5.70 & 15.40$\pm$5.60 & 20.00$\pm$8.40 & 37.90$\pm$12.30 & 25.50$\pm$5.90 & 10.80$\pm$4.50 & 52.20$\pm$9.60 & \textbf{118.19$\pm$7.77} \\
 & random & 6.30$\pm$3.50 & 9.80$\pm$4.90 & 24.00$\pm$9.80 & 34.40$\pm$5.30 & 45.60$\pm$8.70 & 5.50$\pm$1.10 & 62.20$\pm$8.10 & \textbf{130.33$\pm$15.68} \\
\cline{1-10}
\multirow[t]{4}{*}{PP} & expert &  113.00$\pm$14.40* &  115.20$\pm$12.50* & 93.90$\pm$14.00 &  116.20$\pm$19.80* &  108.40$\pm$21.50* &  109.30$\pm$10.10* & \textbf{118.20$\pm$13.10} & Not available \\
 & medium &  63.30$\pm$20.00* &  65.10$\pm$29.50* &  61.70$\pm$23.10* &  66.70$\pm$23.20* &  55.00$\pm$43.20* &  53.60$\pm$19.90* & \textbf{68.50$\pm$21.80} & Not available \\
 & medium-replay & 34.50$\pm$27.80 & 28.70$\pm$20.90 & 24.80$\pm$17.30 & 47.10$\pm$15.30 & 11.90$\pm$9.20 & 23.20$\pm$12.00 & \textbf{71.10$\pm$6.00} & Not available \\
 & random & 2.20$\pm$2.60 & 5.70$\pm$3.50 & 5.00$\pm$8.20 & 11.10$\pm$2.80 & 25.20$\pm$11.50 & 1.30$\pm$1.60 & \textbf{78.50$\pm$15.60} & Not available \\
\cline{1-10}
\multirow[t]{4}{*}{WD} & expert &  109.50$\pm$22.80* &  110.30$\pm$21.30* & 71.90$\pm$28.10 &  110.40$\pm$25.70* &  99.70$\pm$31.00* &  107.80$\pm$17.70* & \textbf{119.70$\pm$26.40} & Not available \\
 & medium &  71.90$\pm$20.00* &  73.40$\pm$9.30* &  58.60$\pm$11.20* &  74.60$\pm$11.50* &  67.40$\pm$48.40* &  70.50$\pm$15.30* & \textbf{93.80$\pm$31.80} & Not available \\
 & medium-replay & 12.00$\pm$9.10 & 17.40$\pm$8.10 & 29.60$\pm$13.80 &  42.90$\pm$19.50* & 13.20$\pm$16.20 & 41.50$\pm$9.50 & \textbf{73.40$\pm$23.20} & Not available \\
 & random & 1.00$\pm$3.20 & 2.80$\pm$5.50 & 0.60$\pm$2.00 & 5.90$\pm$5.20 & 11.70$\pm$11.00 & 2.90$\pm$4.00 & \textbf{68.00$\pm$20.80} & Not available \\

\end{tabular}
\end{adjustbox}
    
    \label{tab:mpe}
\end{table}

% \begin{table}[h!]
% \caption{Results on MPE datasets from OMAR.}
% \begin{adjustbox}{max width=\textwidth}
% \begin{tabular}{ll|rrr}
% Scenario            & Dataset       & MACQL                                 & OMAR        & CFCQL       \\ \hline
% \multirow{4}{*}{CN} & Random        & \result{45.6}{8.7} & 34.4+-5.3   & 62.2+-8.1   \\
%                     & Medium-Replay & 25.5+-5.9                             & 37.9+-12.3  & 52.2+-9.6   \\
%                     & Medium        & 14.3+-20.2                            & 47.9+-18.9  & 65.0+-10.2  \\
%                     & Expert        & 12.2+-31                              & 114.9+-2.6  & 112+-4      \\
% \multirow{4}{*}{PP} & Random        & 25.2+-11.5                            & 11.1+-2.8   & 78.5+-15.6  \\
%                     & Medium-Replay & 11.9+-9.2                             & 47.1+-15.3  & 71.1+-6     \\
%                     & Medium        & 55+-43.2                              & 66.7+-23.2  & 68.5+-21.8  \\
%                     & Expert        & 108.4+-21.5                           & 116.2+-19.8 & 118.2+-13.1 \\
% \multirow{4}{*}{WD} & Random        & 11.7+-11                              & 5.9+-5.2    & 68+-20.8    \\
%                     & Medium-Replay & 13.2+-16.2                            & 42.9+-19.5  & 73.4+-23.2  \\
%                     & Medium        & 67.4+-48.4                            & 74.6+-11.5  & 93.8+-31.8  \\
%                     & Expert        & 99.7+-31                              & 110.4+-25.7 & 119.7+-26.4
% \end{tabular}
% \end{adjustbox}
% \end{table}

\subsection{MAMuJoCo}
\citet{cfcql} normalise the episode returns against the mean in the dataset.

\begin{table}[hbtp]
    \centering
    \caption{Results on MAMuJoCo datasets from OMAR.}
    
\begin{adjustbox}{max width=\textwidth} 
\begin{tabular}{ll|llllllll|ll}

 & & \multicolumn{8}{c}{Shao et al.} & \multicolumn{2}{c}{Our Baselines} \\
 & & maicq & matd3 bc & icql & omar & macql & iql & awac & cfcql & iddpg+bc & maddpg+cql \\
task & dataset quality &  &  &  &  &  &  &  &  &  &  \\
\midrule
\multirow[t]{4}{*}{2x3 halfcheetah} & expert & 110.60$\pm$3.30 & 114.40$\pm$3.80 & 64.20$\pm$24.90 & 113.50$\pm$4.30 & 50.10$\pm$20.10 &  115.60$\pm$4.20* & 113.30$\pm$4.10 &  118.50$\pm$4.90* & \textbf{120.02$\pm$1.80} &  119.69$\pm$3.00* \\
 & medium & 73.60$\pm$5.00 & 75.50$\pm$3.70 & 50.40$\pm$10.80 & 80.40$\pm$10.20 & 51.50$\pm$26.70 & 81.30$\pm$3.70 & 71.20$\pm$4.20 & 80.50$\pm$9.60 & 156.86$\pm$4.53 & \textbf{165.21$\pm$10.96} \\
 & medium-replay & 35.60$\pm$2.70 & 27.10$\pm$5.50 & 41.20$\pm$10.10 & 57.70$\pm$5.10 & 37.00$\pm$7.10 & 58.80$\pm$6.80 & 30.90$\pm$1.60 & 59.50$\pm$8.20 & \textbf{719.03$\pm$37.31} & 668.97$\pm$28.57 \\
 & random & 7.40$\pm$0.00 & 7.40$\pm$0.00 & 7.40$\pm$0.00 & 13.50$\pm$7.00 & 5.30$\pm$0.50 & 7.40$\pm$0.00 & 7.30$\pm$0.00 & \textbf{39.70$\pm$4.00} & Not available & Not available \\

\end{tabular}
\end{adjustbox}
    
    \label{tab:mamujoco_cfcql}
\end{table}

% \begin{table}[h!]
% \caption{Results on MAMuJoCo datasets from OMAR.}
% \begin{adjustbox}{max width=\textwidth}
% \begin{tabular}{ll|rrrr|ll}
%                                  &                  & \multicolumn{4}{c|}{\textbf{Algorithms from the Literature}}     & \multicolumn{2}{c}{\textbf{Our Baselines}} \\
% \textbf{Scenario}                & \textbf{Dataset} & \textbf{MACQL} & \textbf{MAICQ} & \textbf{OMAR} & \textbf{CFCQL} & \textbf{IDDPG+BC}   & \textbf{MADDPG+CQL}  \\ \hline
% \multirow{4}{*}{2x3 HalfCheetah} & Random           & 5.3±0.5        & 7.4±0.0        & 13.5±7.0      & 39.7±4.0       & raw = 40            & raw = -40            \\
%                                  & Medium-Replay    & 37.0±7.1       & 35.6±2.7       & 57.7±5.1      & 59.5±8.2       & \result{719.0}{37}^{**}           & \result{669.0}{28.6}^{**}          \\
%                                  & Medium           & 51.5±26.7      & 73.6±5.0       & 80.4±10.2     & 80.5±9.6       & \result{156}{4.5}^{**}            & \result{165.2}{11.0}^{**}          \\
%                                  & Expert           & 50.1±20.1      & 110.6±3.3      & 113.5±4.3     & 118.5±4.9      & \result{120.0}{1.8}^*          & \result{119.7}{3.0}^*          
% \end{tabular}
% \end{adjustbox}
% \end{table}

In ~\citet{omiga} they give the raw episode returns.

\begin{table}[hbtp]
    \centering
    
    \caption{Results on MAMuJoCo datasets from OMIGA}
    
\begin{adjustbox}{max width=\textwidth} 
\begin{tabular}{ll|lllll|ll}

 & & \multicolumn{5}{c}{Wang et al.} & \multicolumn{2}{c}{Our Baselines} \\
 & & mabcq & macql & maicq & omar & omiga & iddpg+bc & maddpg+cql \\
task & dataset quality &  &  &  &  &  &  &  \\
\midrule
\multirow[t]{4}{*}{2x4 ant} & expert & 1317.73$\pm$286.28 &  1042.39$\pm$2021.65* &  2050.00$\pm$11.86* & 312.54$\pm$297.48 & \textbf{2055.46$\pm$1.58} & 1031.34$\pm$57.83 &  2053.35$\pm$12.02* \\
 & medium & 1059.60$\pm$91.22 &  533.90$\pm$1766.42* &  1412.41$\pm$10.93* & -1710.04$\pm$1588.98 & \textbf{1418.44$\pm$5.36} & 365.73$\pm$54.77 & 1395.29$\pm$15.05 \\
 & medium-expert & 1020.89$\pm$242.74 &  800.22$\pm$1621.52* & 1590.18$\pm$85.61 & -2992.80$\pm$6.95 & 1720.33$\pm$110.63 & 181.87$\pm$352.79 & \textbf{1972.17$\pm$129.61} \\
 & medium-replay & 950.77$\pm$48.76 &  234.62$\pm$1618.28* &  1016.68$\pm$53.51* & -2014.20$\pm$844.68 & \textbf{1105.13$\pm$88.87} & 859.27$\pm$66.68 & 951.95$\pm$10.50 \\
\cline{1-9}
\multirow[t]{4}{*}{3x1 hopper} & expert & 77.85$\pm$58.04 & 159.14$\pm$313.83 & 754.74$\pm$806.28 & 2.36$\pm$1.46 & 859.63$\pm$709.47 & \textbf{3553.13$\pm$107.21} & 82.49$\pm$103.59 \\
 & medium & 44.58$\pm$20.62 & 401.27$\pm$199.88 & 501.79$\pm$14.03 & 21.34$\pm$24.90 &  1189.26$\pm$544.30* & 795.35$\pm$29.69 & \textbf{1745.88$\pm$607.83} \\
 & medium-expert & 54.31$\pm$23.66 & 64.82$\pm$123.31 & 355.44$\pm$373.86 & 1.44$\pm$0.86 & 709.00$\pm$595.66 & \textbf{3087.47$\pm$721.82} & 334.36$\pm$214.95 \\
 & medium-replay & 26.53$\pm$24.04 & 31.37$\pm$15.16 & 195.39$\pm$103.61 & 3.30$\pm$3.22 & 774.18$\pm$494.27 & 231.40$\pm$167.45 & \textbf{1738.38$\pm$291.67} \\
\cline{1-9}
\multirow[t]{4}{*}{6x1 halfcheetah} & expert & 2992.71$\pm$629.65 & 1189.54$\pm$1034.49 & 2955.94$\pm$459.19 & -206.73$\pm$161.12 & 3383.61$\pm$552.67 & \textbf{4763.99$\pm$119.56} & -106.72$\pm$316.55 \\
 & medium &  2590.47$\pm$1110.35* & 1011.35$\pm$1016.94 & 2549.27$\pm$96.34 & -265.68$\pm$146.98 & \textbf{3608.13$\pm$237.37} & 2739.01$\pm$51.09 &  3350.97$\pm$132.93* \\
 & medium-expert & \textbf{3543.70$\pm$780.89} & 1194.23$\pm$1081.06 &  2833.99$\pm$420.32* & -253.84$\pm$63.94 &  2948.46$\pm$518.89* &  3400.65$\pm$454.49* & -183.98$\pm$629.66 \\
 & medium-replay & -333.64$\pm$780.89 & 1998.67$\pm$693.92 & 1922.42$\pm$612.87 & -235.42$\pm$154.89 & 2504.70$\pm$83.47 & \textbf{3380.58$\pm$121.08} & 2795.40$\pm$282.13 \\

\end{tabular}
\end{adjustbox}
    \label{tab:mamujoco_omiga
    }
\end{table}

\subsection{SMAC}
For our SMAC benchmark we managed to get datasets from \citet{cfcql} and from~\citet{omiga}. We then implemented independent Q-learners with CQL. Since we used recurrent Q-networks we call our method IDRQN+CQL.

On the datasets from~\citet{omiga}, the episode returns were given. 

\begin{table}[hbtp]
    \centering
    \caption{Results on SMAC datasets from~\citet{omiga}, subsampled from datasets generated by~\citet{madt}.}
    \begin{adjustbox}{max width=\textwidth} 
\begin{tabular}{ll|lllll|l}

 & & \multicolumn{5}{c}{Wang et al.} & Our Baseline \\
 & & mabcq & macql & maicq & omar & omiga & iql+cql \\
task & dataset quality &  &  &  &  &  &  \\
\midrule
\multirow[t]{3}{*}{2c vs 64zg} & good & 19.13$\pm$0.27 &  18.48$\pm$0.95* & 18.82$\pm$0.17 & 17.27$\pm$0.78 &  19.15$\pm$0.32* & \textbf{19.52$\pm$0.26} \\
 & medium &  15.58$\pm$0.37* & 12.82$\pm$1.61 &  15.57$\pm$0.61* & 10.20$\pm$0.20 & \textbf{16.03$\pm$0.19} & 14.89$\pm$0.72 \\
 & poor &  12.46$\pm$0.18* & 10.83$\pm$0.51 &  12.56$\pm$0.18* & 11.33$\pm$0.50 & \textbf{13.02$\pm$0.66} & 11.03$\pm$0.41 \\
\cline{1-8}
\multirow[t]{3}{*}{5m vs 6m} & good & 7.76$\pm$0.15 & 8.08$\pm$0.21 & 7.87$\pm$0.30 & 7.40$\pm$0.63 & 8.25$\pm$0.37 & \textbf{12.36$\pm$1.09} \\
 & medium & 7.58$\pm$0.10 & 7.78$\pm$0.10 & 7.77$\pm$0.30 & 7.08$\pm$0.51 & 7.92$\pm$0.57 & \textbf{12.30$\pm$0.74} \\
 & poor & 7.61$\pm$0.36 & 7.43$\pm$0.10 & 7.26$\pm$0.19 & 7.27$\pm$0.42 & 7.52$\pm$0.21 & \textbf{10.20$\pm$0.75} \\
\cline{1-8}
\multirow[t]{3}{*}{6h vs 8z} & good & 12.19$\pm$0.23 & 10.44$\pm$0.20 & 11.81$\pm$0.12 & 9.85$\pm$0.28 &  12.54$\pm$0.21* & \textbf{12.72$\pm$0.44} \\
 & medium & 11.77$\pm$0.16 & 11.29$\pm$0.29 & 11.13$\pm$0.33 & 10.36$\pm$0.16 & \textbf{12.19$\pm$0.22} &  12.01$\pm$0.42* \\
 & poor & 10.84$\pm$0.16 &  10.81$\pm$0.52* & 10.55$\pm$0.10 & 10.63$\pm$0.25 & \textbf{11.31$\pm$0.19} & 10.41$\pm$0.36 \\
\cline{1-8}
\multirow[t]{3}{*}{corridor} & good & 15.24$\pm$1.21 & 5.22$\pm$0.81 & 15.54$\pm$1.12 & 6.74$\pm$0.69 & 15.88$\pm$0.89 & \textbf{19.06$\pm$0.81} \\
 & medium & 10.82$\pm$0.92 & 7.04$\pm$0.66 & 11.30$\pm$1.57 & 7.26$\pm$0.71 & 11.66$\pm$1.30 & \textbf{13.44$\pm$1.31} \\
 & poor & 4.47$\pm$0.94 & 4.08$\pm$0.60 & 4.47$\pm$0.33 & 4.28$\pm$0.49 &  5.61$\pm$0.35* & \textbf{6.11$\pm$1.10} \\

\end{tabular}
\end{adjustbox}
    \label{tab:smac_omiga}
\end{table}

On the~\citet{cfcql} datasets, the win rates were given.

\begin{table}[hbtp]
    \centering
    \caption{Results on SMAC datasets from~\citet{cfcql}.}
    \begin{adjustbox}{max width=\textwidth} 
\begin{tabular}{ll|llllllll|l}

 & & \multicolumn{8}{c}{Shao et al.} & Our Baseline \\
 & & cfcql & macql & maicq & omar & madtkd & mabcq & iql & awac & iql+cql \\
task & dataset quality &  &  &  &  &  &  &  &  &  \\
\midrule
\multirow[t]{4}{*}{2s3z} & expert & \textbf{0.99$\pm$0.01} &  0.58$\pm$0.34* & 0.93$\pm$0.04 &  0.95$\pm$0.04* & \textbf{0.99$\pm$0.02} &  0.97$\pm$0.02* &  0.98$\pm$0.03* &  0.97$\pm$0.03* & \textbf{0.99$\pm$0.03} \\
 & medium & \textbf{0.40$\pm$0.10} & 0.17$\pm$0.08 & 0.18$\pm$0.02 & 0.15$\pm$0.04 & 0.18$\pm$0.03 & 0.16$\pm$0.07 & 0.16$\pm$0.04 & 0.19$\pm$0.05 & 0.16$\pm$0.13 \\
 & medium-replay & \textbf{0.55$\pm$0.07} & 0.12$\pm$0.08 & 0.41$\pm$0.06 & 0.24$\pm$0.09 & 0.36$\pm$0.07 & 0.33$\pm$0.04 & 0.33$\pm$0.06 & 0.39$\pm$0.05 & 0.33$\pm$0.08 \\
 & mixed & 0.84$\pm$0.09 & 0.67$\pm$0.17 & 0.85$\pm$0.07 & 0.60$\pm$0.04 & 0.47$\pm$0.08 & 0.44$\pm$0.06 & 0.19$\pm$0.04 & 0.14$\pm$0.04 & \textbf{0.97$\pm$0.03} \\
\cline{1-11}
\multirow[t]{4}{*}{3s vs 5z} & expert & \textbf{0.99$\pm$0.01} & 0.92$\pm$0.05 & 0.91$\pm$0.04 & 0.64$\pm$0.08 & 0.67$\pm$0.08 &  0.98$\pm$0.02* & \textbf{0.99$\pm$0.01} & \textbf{0.99$\pm$0.02} &  0.98$\pm$0.03* \\
 & medium & \textbf{0.28$\pm$0.03} & 0.09$\pm$0.06 & 0.03$\pm$0.01 & 0.00$\pm$0.00 & 0.01$\pm$0.01 & 0.08$\pm$0.02 & 0.20$\pm$0.05 & 0.19$\pm$0.03 & 0.13$\pm$0.07 \\
 & medium-replay & 0.12$\pm$0.04 & 0.01$\pm$0.01 & 0.01$\pm$0.02 & 0.00$\pm$0.00 & 0.01$\pm$0.01 & 0.01$\pm$0.01 & 0.04$\pm$0.04 & 0.08$\pm$0.05 & \textbf{0.26$\pm$0.16} \\
 & mixed &  0.60$\pm$0.14* & 0.17$\pm$0.10 & 0.10$\pm$0.04 & 0.00$\pm$0.00 & 0.14$\pm$0.08 & 0.21$\pm$0.04 & 0.20$\pm$0.06 & 0.18$\pm$0.03 & \textbf{0.67$\pm$0.18} \\
\cline{1-11}
\multirow[t]{4}{*}{5m vs 6m} & expert & \textbf{0.84$\pm$0.03} & 0.01$\pm$0.01 & 0.72$\pm$0.05 & 0.33$\pm$0.06 & 0.58$\pm$0.07 &  0.82$\pm$0.04* & 0.77$\pm$0.03 & 0.75$\pm$0.02 & 0.72$\pm$0.11 \\
 & medium & \textbf{0.29$\pm$0.05} & 0.01$\pm$0.01 &  0.26$\pm$0.03* & 0.19$\pm$0.06 & 0.21$\pm$0.04 &  0.28$\pm$0.37* &  0.25$\pm$0.02* & 0.22$\pm$0.04 & 0.19$\pm$0.09 \\
 & medium-replay & \textbf{0.22$\pm$0.06} &  0.16$\pm$0.08* &  0.18$\pm$0.04* & 0.03$\pm$0.02 &  0.16$\pm$0.04* &  0.18$\pm$0.06* &  0.18$\pm$0.04* &  0.18$\pm$0.04* & \textbf{0.22$\pm$0.16} \\
 & mixed &  0.76$\pm$0.07* & 0.01$\pm$0.01 & 0.67$\pm$0.08 & 0.10$\pm$0.10 & 0.21$\pm$0.05 & 0.21$\pm$0.12 &  0.76$\pm$0.06* & \textbf{0.78$\pm$0.02} & 0.65$\pm$0.16 \\
\cline{1-11}
\multirow[t]{4}{*}{6h vs 8z} & expert & \textbf{0.70$\pm$0.06} & 0.00$\pm$0.00 & 0.24$\pm$0.08 & 0.01$\pm$0.01 & 0.48$\pm$0.08 & 0.60$\pm$0.04 &  0.67$\pm$0.03* &  0.67$\pm$0.03* &  0.60$\pm$0.17* \\
 & medium &  0.41$\pm$0.04* & 0.01$\pm$0.01 & 0.19$\pm$0.04 & 0.04$\pm$0.03 & 0.22$\pm$0.07 &  0.40$\pm$0.03* &  0.40$\pm$0.05* & \textbf{0.43$\pm$0.06} &  0.32$\pm$0.17* \\
 & medium-replay &  0.21$\pm$0.05* & 0.08$\pm$0.04 & 0.07$\pm$0.04 & 0.00$\pm$0.00 &  0.12$\pm$0.05* & 0.11$\pm$0.04 &  0.17$\pm$0.03* &  0.14$\pm$0.04* & \textbf{0.22$\pm$0.13} \\
 & mixed & 0.49$\pm$0.08 & 0.01$\pm$0.01 & 0.05$\pm$0.03 & 0.00$\pm$0.00 & 0.25$\pm$0.07 & 0.27$\pm$0.06 & 0.36$\pm$0.05 & 0.35$\pm$0.06 & \textbf{0.63$\pm$0.12} \\

\end{tabular}
\end{adjustbox}
    \label{tab:smac_cfcql}
\end{table}

\newpage
\section{Converted Datasets}

\begin{mybox}{\small Machine Learning Reproducibility Checklist (Datasets)}{box:mlcheck_datasets}
\small
\begin{enumerate}
\item For all datasets used, check if you include:
\begin{enumerate}
  \item The relevant statistics, such as number of examples.
    \answerYes{See provided notebook and sample code below for analysing datasets.}
  \item The details of train / validation / test splits.
    \answerNo{Not applicable for offline MARL in this case.}
  \item An explanation of any data that were excluded, and all pre-processing step.
    \answerYes{A full explanation in the main body of the text and below; we converted the datasets to the vault API.}
  \item A link to a downloadable version of the dataset or simulation environment.
    \answerYes{See below.}
  \item For new data collected, a complete description of the data collection process, such as
instructions to annotators and methods for quality control.
    \answerNo{Not applicable; all datasets used are from the literature.}
\end{enumerate}

\end{enumerate}
\end{mybox}

To compare existing results to our own baselines, we converted datasets from the literature into the Vault format from \href{https://github.com/instadeepai/flashbax}{Flashbax}. Examples of our methodology for converting these datasets are given in a notebook here: \url{https://bit.ly/vault-conversion-notebook}. Table~\ref{tab:vault_links} provides links to all Vault datasets.

\begin{table}[h]
\centering
\scriptsize
\setstretch{1.5}
\caption{Links to Vault datasets converted from the literature.}\label{tab:vault_links}
\vspace{1em}
\begin{adjustbox}{max width=\textwidth}
\begin{tabular}{l|lcl}
\textbf{Paper}& \textbf{Environment} & \textbf{Scenario} & \textbf{Link} \\ \hline
% \cite{formanek2023og-marl} & SMACv1 & \texttt{3m} & \url{https://huggingface.co/datasets/InstaDeepAI/og-marl/resolve/main/og_marl/smac_v1/3m.zip} \\ 
%  &  & \texttt{8m} & \url{https://huggingface.co/datasets/InstaDeepAI/og-marl/resolve/main/og_marl/smac_v1/8m.zip} \\ 
%  &  & \texttt{5m\_vs\_6m} & \url{https://huggingface.co/datasets/InstaDeepAI/og-marl/resolve/main/og_marl/smac_v1/5m_vs_6m.zip} \\ 
%  &  & \texttt{2s3z} & \url{https://huggingface.co/datasets/InstaDeepAI/og-marl/resolve/main/og_marl/smac_v1/2s3z.zip} \\ 
%  &  & \texttt{3s5z\_vs\_3s6z} & \url{https://huggingface.co/datasets/InstaDeepAI/og-marl/resolve/main/og_marl/smac_v1/3s5z_vs_3s6z.zip} \\ 
%  &  & \texttt{2c\_vs\_64zg} & \url{ } \\
%  & MAMuJoCo & \texttt{2ant} & \url{ } \\
%  &  & \texttt{2halfcheetah} & \url{https://huggingface.co/datasets/InstaDeepAI/og-marl/resolve/main/og_marl/mamujoco/2halfcheetah.zip} \\
%  &  & \texttt{4ant} & \url{ } \\ \hline
\cite{omar} & MAMuJoCo & \texttt{2halfcheetah} & \url{https://huggingface.co/datasets/InstaDeepAI/og-marl/resolve/main/prior_work/omar/mamujoco/2halfcheetah.zip} \\
 & MPE & \texttt{simple-spread} & \url{https://huggingface.co/datasets/InstaDeepAI/og-marl/resolve/main/prior_work/omar/mpe/simple_spread.zip} \\
 & & \texttt{simple-tag} & \url{https://huggingface.co/datasets/InstaDeepAI/og-marl/resolve/main/prior_work/omar/mpe/simple_tag.zip} \\
 & & \texttt{simple-world} & \url{https://huggingface.co/datasets/InstaDeepAI/og-marl/resolve/main/prior_work/omar/mpe/simple_world.zip} \\\hline
\cite{cfcql} & SMACv1 & \texttt{2s3z} & \url{https://huggingface.co/datasets/InstaDeepAI/og-marl/resolve/main/prior_work/cfcql/smac_v1/2s3z.zip} \\
 &  & \texttt{3s\_vs\_5z} & \url{https://huggingface.co/datasets/InstaDeepAI/og-marl/resolve/main/prior_work/cfcql/smac_v1/3s_vs_5z.zip} \\
 &  & \texttt{5m\_vs\_6m} & \url{https://huggingface.co/datasets/InstaDeepAI/og-marl/resolve/main/prior_work/cfcql/smac_v1/5m_vs_6m.zip} \\
 &  & \texttt{6h\_vs\_8z} & \url{https://huggingface.co/datasets/InstaDeepAI/og-marl/resolve/main/prior_work/cfcql/smac_v1/6h_vs_8z.zip} \\\hline
\cite{omiga} & SMACv1 & \texttt{corridor} & \url{https://huggingface.co/datasets/InstaDeepAI/og-marl/resolve/main/prior_work/omiga/smac_v1/corridor.zip} \\
 &  & \texttt{2c\_vs\_64zg} & \url{https://huggingface.co/datasets/InstaDeepAI/og-marl/resolve/main/prior_work/omiga/smac_v1/2c_vs_64zg.zip} \\
 &  & \texttt{5m\_vs\_6m} & \url{https://huggingface.co/datasets/InstaDeepAI/og-marl/resolve/main/prior_work/omiga/smac_v1/5m_vs_6m.zip} \\
 &  & \texttt{6h\_vs\_8z} & \url{https://huggingface.co/datasets/InstaDeepAI/og-marl/resolve/main/prior_work/omiga/smac_v1/6h_vs_8z.zip} \\
 & MAMuJoCo & \texttt{2ant} & \url{https://huggingface.co/datasets/InstaDeepAI/og-marl/resolve/main/prior_work/omiga/mamujoco/2ant.zip} \\
 & & \texttt{3hopper} & \url{https://huggingface.co/datasets/InstaDeepAI/og-marl/resolve/main/prior_work/omiga/mamujoco/3hopper.zip} \\
 & & \texttt{6halfcheetah} & \url{https://huggingface.co/datasets/InstaDeepAI/og-marl/resolve/main/prior_work/omiga/mamujoco/6halfcheetah.zip} \\
\end{tabular}
\end{adjustbox}
\end{table}

% \newpage
\paragraph{Statistical profiles}
The statistical profiles (histograms and violin plots) of the converted datasets are available in the respective folders on HuggingFace: \url{https://huggingface.co/datasets/InstaDeepAI/og-marl/tree/main/prior_work}. Additionally, summary statistic values can easily be accessed using the demonstrative notebook at: \url{https://github.com/instadeepai/og-marl/blob/main/examples/dataset_analysis_demo.ipynb}.

\end{document}